\documentclass{article} 
\usepackage{iclr2027_conference,times}

\usepackage{amsmath,amsfonts,bm}

\def\eqref#1{equation~\ref{#1}}

\def\1{\bm{1}}

\DeclareMathAlphabet{\mathsfit}{\encodingdefault}{\sfdefault}{m}{sl}
\SetMathAlphabet{\mathsfit}{bold}{\encodingdefault}{\sfdefault}{bx}{n}

\usepackage{hyperref}
\usepackage{url}

\usepackage{graphicx}
\usepackage{booktabs}
\usepackage{array}
\usepackage{capt-of}

\title{Objects as Addresses: Instance Binding in Vision Transformer Attention}

\author{
\begin{minipage}{0.95\textwidth}
\centering
Khalil Sabri$^{1}$ \qquad
Guillaume-Alexandre Bilodeau$^{1}$\\
Nicolas Saunier$^{1}$ \qquad
Wassim Bouachir$^{2}$\\[4pt]
{\normalfont\small
$^{1}$Polytechnique Montr\'eal \quad
$^{2}$Universit\'e T\'ELUQ\\
Montr\'eal, Canada}\\[3pt]
{\normalfont\footnotesize
\texttt{\{khalil.sabri,gabilodeau,nicolas.saunier\}@polymtl.ca}\\
\texttt{wassim.bouachir@teluq.ca}}
\end{minipage}
}

\iclrfinalcopy 
\begin{document}

\maketitle


\begin{abstract}
Humans naturally parse visual scenes into coherent individual objects rather than scattered features, a capability cognitive science calls object binding. While recent work shows that pretrained Vision Transformers (ViTs) intrinsically acquire this ability, how native self-attention computations mechanistically keep object instances separate remains unresolved. By studying this across eight pretrained ViTs, we found that instance separation is expressed more strongly in attention routing than in token content: selected query--key relations distinguish individual objects more reliably than value similarity or residual representations. Inside heads, patches belonging to one individual tend to read from a common image region, which we term their \emph{attention address}. Visually distinct parts of one person share an address, whereas identical duplicated instances receive separate, movable addresses. These shared destinations arise in part because query interactions with the spatial structure of keys direct attention toward consistent regions for one object and elsewhere for another. In DINOv2 and most other backbones we tested, editing this query component at the selected head redirects its attention address toward another individual. Grouping patches by this query component separates same-category instances, reaching .615 FG-ARI versus .289 for the same clustering of final-layer features, which tend to merge them.
\end{abstract}

\begin{center}
\begin{minipage}{\textwidth}
\centering
\includegraphics[width=0.75\textwidth]{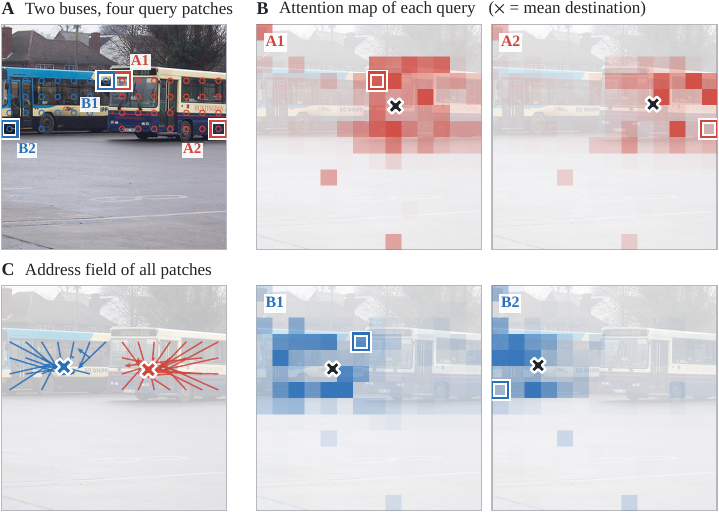}
\captionof{figure}{\textbf{Objects as addresses.} The cross marks the mean destination of each map. Colors use ground truth only to show which patches belong together. (A) Four patches query the same frozen attention head, two on each bus. (B) Patches on one bus can look different, yet their complete attention maps read from a common region.  (C) Drawing an arrow from every bus patch to its address turns the head into an address field, in which the patches of each bus point to one region. }
\label{fig:teaser}

\end{minipage}
\end{center}

\section{Introduction}
\label{sec:intro}

An image can contain two zebras. However, labeling every relevant patch as \emph{zebra} is not enough: the model must still associate each head, torso, and leg with the correct individual. This is the visual binding problem, representing not only \emph{what} features are present, but \emph{which features belong together} \citep{treisman1980feature,greff2020binding}. This problem was tackled by \citet{li2025binding}, who recovered patch-level same-object membership, \textsc{IsSameObject}, with a learned quadratic probe and connected it to attention. In contrast, our aim is to identify how existing ViT attention distinguishes individuals, then to explain and manipulate the part of its computation that directs where their patches read. We focus on objects of the same category. Nearby patches often belong to the same object, as do patches with similar semantics, and either cue can obscure whether an attention head distinguishes individuals. We therefore compare same-instance pairs of patches with pairs from different instances of one category in the same image and at exactly the same relative offset (same distance and direction). We call this controlled problem \emph{object individuation}.

We used Pascal VOC for development and held-out evaluation, and COCO for transfer and later analyses. Labels define the test and are used to select the layers, heads, and relations on the VOC development split. Evaluation uses disjoint held-out VOC images. COCO is not used for component selection unless stated otherwise. No binding predictor is trained.

By analyzing the results, we identified heads whose reading distinguishes individuals better than residual or value similarity. Following their attention reveals a common destination for patches of one object. We then examined how a query interaction with spatial structure in the keys helps produce that destination.
Our contributions are:

\begin{enumerate}\setlength\itemsep{2pt}
  \item \textbf{A controlled test that localizes instance-sensitive attention.} Across eight pretrained ViTs, query/key comparisons and attention separate individuals more strongly than value or residual similarity. This advantage concentrates in particular heads and layers.
  \item \textbf{A shared reading destination for parts of one object.} Patches of an individual tend to read from a common region, their \emph{attention address}. Identical copies receive different addresses, and an address moves with its object.
  \item \textbf{An interpretable spatial request within attention.} A query interaction with the keys' spatial structure specifies where it would read, its \emph{spatial request}. This request retains instance discrimination across several backbones. In DINOv2, its destination predicts the attention address, and editing it redirects the head toward another individual.
\end{enumerate}

\section{Related work}
\label{sec:related}

\paragraph{Binding and object representations.}
The binding problem asks how visual features are organized into coherent objects, rather than represented independently \citep{treisman1980feature,pylyshyn1989finst,kahneman1992object,greff2020binding}. \citet{li2025binding} decoded same-object membership from frozen ViTs with a quadratic probe and linked this signal to attention. \citet{huang2026formalizing} formalized binding information, that is, which features belong to the same object, and measured it in ViT representations, while \citet{tartaglini2026continuity} found mostly early heads that track continuity in synthetic stimuli and contribute to binding. We ask a more specific question: which native attention computations keep individuals of the same category apart, and how?

\paragraph{Object-centric learning and pretrained ViTs.}
Object-centric learning makes object assignment explicit through representations such as slots \citep{locatello2020slot,seitzer2023dinosaur}. Yet instance identity remains challenging: SemanticSlots \citep{semanticslots} identified ambiguity between identical entities as a limitation of semantic retrieval, while SlotContrast \citep{manasyan2025slotcontrast} used temporal consistency to preserve object identity across frames. Decoder comparisons in DINOSAUR-style systems \citep{seitzer2023dinosaur} also showed that the decoding choice can substantially change instance grouping. This motivates asking what instance information is already present in the pretrained ViT before an object-centric module decides what to preserve. Self-supervised ViTs already expose object structure through attention and feature affinities \citep{caron2021dino,simeoni2021lost,wang2022tokencut,melaskyriazi2022deep}, while Object-DINO \citep{rawlekar2026objectdino} finds complementary object information across Q, K, and V.

\paragraph{Attention, space, and binding.}
\citet{zhang2024whatwhere} found that Q/K carries strong spatial structure while V is more semantic. \citet{pan2024dissecting} decomposed the native Q/K interaction and found that early layers attend more to similar tokens and late layers to dissimilar, contextual ones. Prior work also studied spatial structure in attention \citep{cordonnier2020relationship,dascoli2021convit,kim2023gaussian}. We examine how a pretrained head uses that structure to distinguish instances. Related mechanisms appear in language and VLMs: binding IDs link entities to attributes \citep{feng2023binding}, while spatial IDs let VLMs retrieve particular visual objects \citep{assouel2026symbolic} and reason about their locations \citep{kang2026spatiotemporal}. We study this inside the frozen visual encoder itself: which native attention components distinguish individual objects, where those objects are read from, and whether editing the query component that sets this reading redirects the head to another instance.

\section{Measuring object individuation in native attention}
\label{sec:measurement}

We first compare the different parts of attention to determine where the same-instance signal appears.

\subsection{Where to read and what to carry}

\citet{li2025binding} predict whether two patches $h_i$ and $h_j$ belong to the same object with a learned bilinear score (probe) $h_i^\top W h_j$, where $W$ is a probe matrix learned from same-object labels. However, a ViT already compares patches through a related score. In each attention head \citep{vaswani2017attention}, the query of a patch is compared with the keys of all patches to decide where it reads, and a weighted average of their values is what it retrieves. \citet{elhage2021framework} described these two roles as a query--key circuit, which computes the attention pattern, and an output--value circuit, which decides what is moved. For one attention head, let $h_i$ and $h_j$ be the normalized input representations of patches $i$ and $j$, let $q$, $k$, and $v$ denote queries, keys, and values, and let $W_Q$, $W_K$, and $W_V$ be their projection matrices. Let $d$ be the query--key dimension. Suppressing affine biases,
\begin{equation}
\begin{aligned}
q_i&=W_Qh_i,\quad k_i=W_Kh_i,\quad v_i=W_Vh_i,\\
q_i^\top k_j&=h_i^\top W_Q^\top W_Kh_j,\\
A_{ij}&=\operatorname{softmax}_j(q_i^\top k_j/\sqrt d),
\end{aligned}
\label{eq:attention}
\end{equation}
where the softmax normalizes over destinations $j$, and the head returns $o_i=\sum_j A_{ij}v_j$. The probe of \citet{li2025binding} learns $W$ from object labels. The head already has its own learned comparison, $W_Q^\top W_K$, which we can test directly. Similarity between two queries (Q--Q) or two keys (K--K) offers another way to test the same head. We therefore ask: which representations and operations of the frozen head expose the same-instance relation?

To answer this, we compared patches through their residual representations (H--H), queries (Q--Q), keys (K--K), and values (V--V), using cosine similarity. We also compared the query--key scores ($Q{\leftrightarrow}K$) and attention weights of the head, averaging both directions because same-instance membership is symmetric. Appendix~\ref{app:protocol} gives the scoring details.
We call query similarity, key similarity, the query--key relation, and attention the \emph{routing-side} (Q/K) relations.

\subsection{A controlled test of object individuation}
\label{sec:controlledtest}

\begin{figure}[t]
  \centering
  \includegraphics[width=\textwidth]{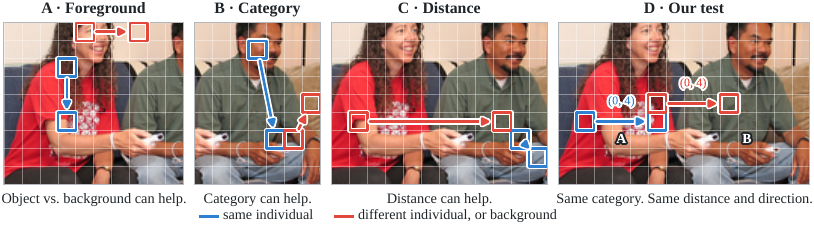}
  \caption{\textbf{Testing which individual, beyond easier cues.} In an unrestricted patch-pair task, object--background differences (A), category differences (B), or unequal distances (C) can help distinguish pair labels. Our test compares patches on objects of the same category and balances same- and different-instance pairs at every offset, such as the $(0,+4)$ shared by the two arrows in D.}
  \label{fig:taskcontrols}
\end{figure}

We used the following experimental setup. A positive pair contains two patches from one annotated instance, while a negative pair contains patches from two different instances of the same category. All patches belong to foreground objects in the same image, and positive and negative pairs occur equally often at every relative offset, including its direction. For example, if two patches of zebra A are separated by $(+3,+1)$, the comparison includes equally many cross-instance pairs with that offset. By doing this, category agreement, foreground membership, distance, and direction cannot solve the task. Each relation scores every pair, and we report the AUC of ranking positive above negative pairs, per image. Without offset matching, proximity alone reaches $.79$ AUC on same-category COCO pairs, and matching reduces it to chance. Figure~\ref{fig:taskcontrols} illustrates the cues our controlled test removes.

We evaluated frozen ViT-B backbones trained with DINO, DINOv2, DINOv2 with registers, DINOv3, MAE, CLIP, SigLIP, and supervised ImageNet classification. We split VOC into a development set, used to select components, and a disjoint held-out set used for evaluation. For each relation, a layer and head are chosen on the development split and then applied unchanged to held-out VOC and to COCO. Appendix~\ref{app:protocol} gives the exact protocol.

\begin{figure}[t]
  \centering
  \includegraphics[width=\textwidth]{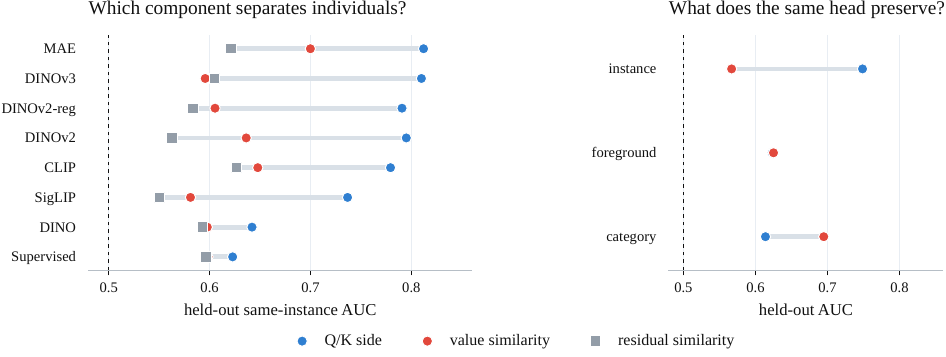}
  \caption{\textbf{Instance-sensitive heads separate routing from content.} Left: Q/K, value and residual comparisons each use their own development-selected component. Q/K separates individuals best in every backbone. Right: the same heads on two further pair tests, foreground versus background patches and one category versus another (Appendix~\ref{app:protocol}). Q/K better preserves instance membership while values better preserve category. Foreground-pair scores are similar on average.}
  \label{fig:qkv}
\end{figure}

\subsection{Routing distinguishes individuals while values preserve category}
\label{sec:routing}

Results show that the routing side separates individuals better than value or residual similarity in every backbone (Fig.~\ref{fig:qkv}, left). The complete breakdown confirms that query similarity, key similarity, the query--key relation and attention all expose individuation more strongly than residual or value similarity (Table~\ref{tab:operands}). Head rankings transfer from VOC to COCO (Spearman $.93$--$.98$).

Figure~\ref{fig:qkv}(right) reveals a more pronounced division. At the same instance-sensitive heads, the routing side is the strongest for \emph{which individual} a patch belongs to, while V is the strongest for \emph{what category} that individual belongs to. This goes beyond a generic ``where versus what'' distinction \citep{zhang2024whatwhere}: relative position is matched, yet the routing side still keeps zebra A separate from zebra B while the values preserve that both are zebras.

Q, K, and V are projections of the same residual state, so this separation is not automatic. Residual similarity reaches only $.55$--$.63$ AUC across models, compared with $.62$--$.81$ for the best held-out routing-side relation. The instance signal is therefore much easier to see inside selected attention heads than in ordinary residual similarity. The instance signal concentrates in particular heads and layers. In most backbones, a small set of heads stands out and the rest are much weaker (Appendix~\ref{app:qk}). In DINOv2, DINOv3 and MAE the strongest heads lie in the last layers. In the supervised model, no head stands out. This localization is why we study one DINOv2 head in detail below.

\section{Following where instance-sensitive heads read}
\label{sec:addresses}

What does an instance-sensitive head actually do? We now follow where each patch reads, with the same experimental setup as in Section~\ref{sec:measurement} and the heads selected there kept frozen. We compare full attention patterns, summarize each pattern by one destination, and test what this destination tracks.

\subsection{Shared reading and the attention address}
\label{sec:fullpattern}

Direct attention considers whether two patches attend to one another. The \emph{full attention pattern} of a patch is its row of $A$, its weights over all destinations. Do patches from the same individual \emph{read the rest of the scene in the same way}? The head and leg of zebra A can look less alike than the heads of zebras A and B, yet both parts of zebra A can place attention on the same regions. Similarity of full attention patterns distinguishes same-instance pairs better than direct attention in nearly every model and dataset we tested (Appendix~\ref{app:sources}, Table~\ref{tab:overlap}). \emph{Reading from similar destinations is a stronger same-instance signal than attending to one another.}

For each patch, we summarize where its attention lands by its mean destination. We call this its \emph{attention address}. Let $p_j$ be the (row, column) position of patch $j$ on the $14\times14$ grid. Attention can also land on prefix or register tokens \citep{darcet2024registers}, which have no image position, so we renormalize it over the image patches $\mathcal P$: $\widetilde A_{ij}=A_{ij}/\sum_{k\in\mathcal P}A_{ik}$. The address of patch $i$ is
\begin{equation}
\mu_i=\sum_{j\in\mathcal P}\widetilde A_{ij}p_j.
\label{eq:address}
\end{equation}
Drawing an arrow from $p_i$ to $\mu_i$ for every patch turns the head into a visible \emph{address field} (Figs.~\ref{fig:teaser} and~\ref{fig:copies}). Patches from one object often point toward one region, while patches from another point elsewhere. The address is not necessarily the object center or the location of a sharp attention peak. It is the mean destination of the full attention pattern, which may contain several peaks.

\subsection{The address follows the individual, not its appearance}

If the address belongs to an individual rather than to its appearance, different-looking parts of one individual should share it, and identical copies should not. We tested these hypotheses in DINOv2.

\paragraph{Different-looking parts share an address.} In COCO, images containing several people, two different body parts of the same person, such as a head and a torso, form a positive pair. The same two parts on different people form a negative, with displacement matched as before (Appendix~\ref{app:crossparts}). Despite their different appearance, the addresses distinguish individuals at $.935$ AUC, close to the full attention pattern and well above residual, value, and query similarity (Table~\ref{tab:crossparts}, right).

\begin{figure}[t]
  \centering
  \includegraphics[width=\textwidth]{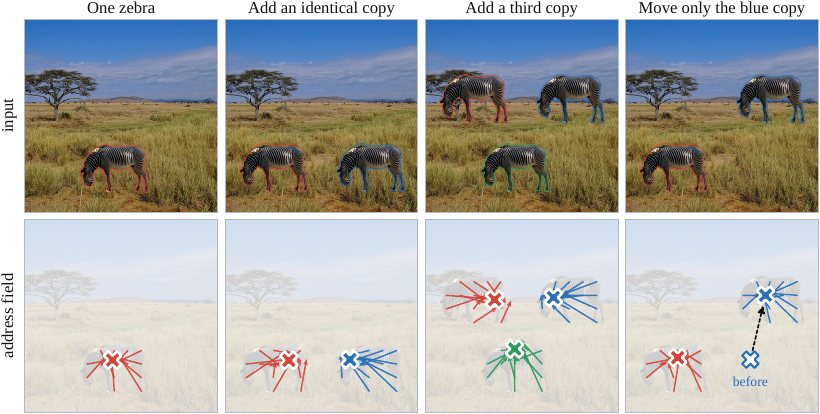}
  \caption{\textbf{Identical appearance, separate addresses.} Pixel-identical copies of one zebra are placed on the same background (top). The patches of each copy point to their own destination (bottom; crosses mark the mean address of each copy). When only the blue copy moves, its address follows it, while the address of the red copy barely changes.}
  \label{fig:copies}
\end{figure}

\paragraph{Identical copies receive different addresses.} If the address merely described appearance, identical copies would collide. We place pixel-identical copies of a segmented zebra on a natural background. Each copy receives its own address $\mu$. When one copy moves, its address follows it, while the address of the stationary copy barely changes (Fig.~\ref{fig:copies}). \citet{li2025binding} find that their probe separates identical objects in middle layers but not near the top of the network. Here, late-layer heads keep the copies apart, at layer 10 of 12 in both DINOv2 and DINOv3 (Appendix~\ref{app:copies}).

The address thus belongs to the individual rather than to its appearance. Why does the head produce such shared destinations for each instance?

\section{How a query specifies where to read}
\label{sec:reverse}

The attention address summarizes where attention lands once the head has compared a query with every key in the image. It shows where a patch reads, not how the query--key computation chose that destination. \emph{How can patches that look different end up reading from the same region?} Each query compares against a key at every image location. One simple hypothesis is that, for a patch on zebra A, these scores rise toward one region of the image and fall away from it, like a smooth peak, whatever that region contains. Scores of this kind would arise if part of each key encoded where its patch lies, and the query weighted this part to select a region. We first look for such a spatial pattern in the keys, then describe how each query uses it (Section~\ref{sec:construction}). Because this description belongs to the query, we can compare it with the observed address (Section~\ref{sec:geometrytransfer}) and edit it (Section~\ref{sec:whatitdoes}).

We focus our study on the DINOv2 \citep{oquab2023dinov2} head selected by the controlled test of Section~\ref{sec:controlledtest}, layer 10, head 2 (L10h2), with the same development and held-out splits. Each patch on its $14\times14$ grid has a 64-dimensional query and key.

\subsection{From a spatial pattern in the keys to a spatial request}
\label{sec:construction}

\paragraph{Keys contain a simple spatial pattern.} A key depends on both what is in the image and where its patch lies. To see what stays tied to location, we averaged keys from the same patch position across development images. Differences between images are reduced, while patterns that recur at the same positions remain. Content that often appears in one part of the image also affects this average. We therefore examine only whether a simple function of position describes the average key well. The form of this function matters because a query scores each location through the dot product with the key there: for a patch on zebra A to favor zebra A, these products must be higher on zebra A than elsewhere. A key that varies linearly with position lets a query favor only one side of the image. A quadratic form can also favor a region between the edges, such as one zebra, and is the lowest order that can do so. After the softmax, a quadratic score can form a Gaussian-shaped peak of attention anywhere in the image. With $r$ and $c$ the row and column coordinates of the patch grid, centered and scaled, the average key at $(r,c)$ is
\begin{equation}
\bar k(r,c)\approx b_0+r\,b_r+c\,b_c+r^2b_{r^2}+c^2b_{c^2}+rc\,b_{rc}.
\label{eq:scaffold}
\end{equation}
Each coefficient $b$ is a 64-dimensional vector, fitted by least squares from patch coordinates and average keys only, then kept fixed. No object labels are used, the ViT is unchanged, and no new module is added. We call these vectors the \emph{spatial basis}. The quadratic function captures most, but not all, of the variation of the mean key across positions: $91\%$ on the development split, against $67\%$ for a linear fit. A cubic fits slightly better without a consistent gain in routing or grouping (Appendix~\ref{app:theta}, Table~\ref{tab:curvature}).

\begin{figure}[t]
  \centering
  \includegraphics[width=\textwidth]{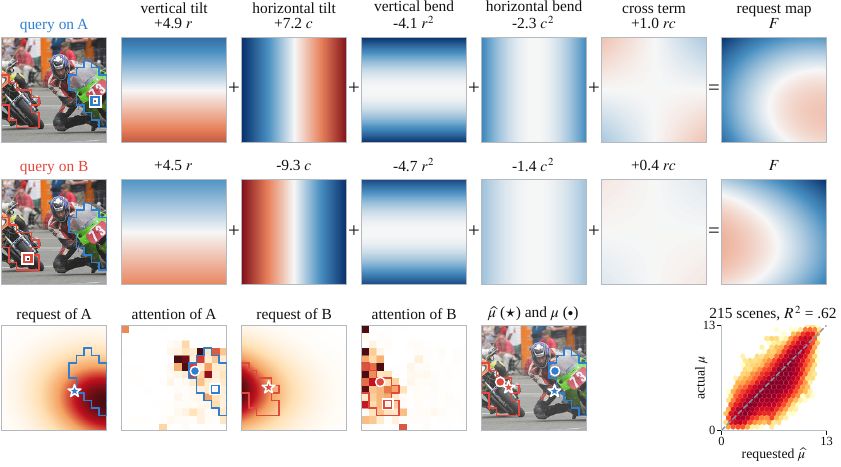}
  \caption{\textbf{A query builds its spatial request from five position maps.} Top rows: for one query on each motorcyclist, the request $\theta$ weights the five position maps of the spatial basis, and their sum is the request map $F$ (Eq.~\ref{eq:field}). Bottom: the softmax of $F$ gives the request distribution $\widehat A$, whose mean is the requested destination $\widehat\mu$ ($\star$). The actual attention of the same query reads nearby, with address $\mu$ ($\bullet$). Bottom right: across all patches of 215 COCO scenes, requested and actual destinations track one another. The dashed line is exact agreement.}
  \label{fig:theta}
\end{figure}

\paragraph{The query weighs this pattern.} If we replace each key by its spatial description, the score of query $i$ at location $(r,c)$ becomes $q_i^\top b_0+r\,q_i^\top b_r+c\,q_i^\top b_c+\dots$. The first term is the same for every destination and cancels in the softmax. The query thus interacts with the spatial pattern through five coefficients. Let $B$ be the $5\times64$ matrix whose rows are the five nonconstant basis vectors. We call these five coefficients, $\theta_i=Bq_i$, the \emph{spatial request} of patch $i$, or its request for short. They define the \emph{request map} $F_i$, the score that the query gives each location through the spatial basis alone:
\begin{equation}
\begin{aligned}
\theta_i&=(\alpha_i,\beta_i,\gamma_i,\delta_i,\epsilon_i)=Bq_i,\\
F_i(r,c)&=\alpha_i r+\beta_i c+\gamma_i r^2+\delta_i c^2+\epsilon_i rc.
\end{aligned}
\label{eq:field}
\end{equation}
The two linear terms tilt this map across rows and columns, and the other three bend it (Fig.~\ref{fig:theta}). The softmax of the head turns $F_i$ into a distribution over image patches, $\widehat A_i$. Intuitively, $\widehat A_i$ shows where patch $i$ would read if the keys carried only their spatial pattern: a smooth region, which in Fig.~\ref{fig:theta} lies on the rider of each query. We summarize it like the address, by its mean location, the \emph{requested destination} $\widehat\mu_i$:
\begin{equation}
\widehat A_{ij}=\operatorname{softmax}_{j\in\mathcal P}\big(F_i(r_j,c_j)/\sqrt d\big),\qquad
\widehat\mu_i=\sum_{j\in\mathcal P}\widehat A_{ij}p_j.
\label{eq:request}
\end{equation}

\paragraph{Address and request.} The address $\mu_i$ is measured: where patch $i$ reads in this image. The request $\theta_i$ is computed from the query and the fixed spatial basis: where the query would read if keys carried only their spatial pattern. Up to a constant that cancels in the softmax, every actual attention score is the request score $F_i$ plus a term from key content that the basis omits, such as image-specific details (Appendix~\ref{app:theta}). The request is therefore the spatial part of the native score, not a substitute for attention, and this remainder is why $\widehat\mu_i$ predicts $\mu_i$ only in part.

\subsection{The request retains the instance relation and predicts the address}
\label{sec:geometrytransfer}

If the request is part of how L10h2 routes, it should keep the same-instance relation that selected this head, and its destination should predict where attention lands.

On the matched patch pairs of Section~\ref{sec:controlledtest}, the requested destination $\widehat\mu$ separates individuals. On held-out VOC it reaches $.872$ AUC, above $.795$ for direct attention, and the advantage persists on COCO. It falls close to chance in an untrained network or when requests come from another image (Appendix~\ref{app:theta}). At the selected head of every backbone, the request matches or improves on the original Q/K relation, and the requested destination separates individuals at least as well as the observed address (Table~\ref{tab:transfer}, left). The separation that the address showed in Section~\ref{sec:addresses} is thus already present in this component of the query.

In Figure~\ref{fig:theta}, the requested destination of each query lies near its observed address. Across all patches of 215 COCO scenes, $\widehat\mu$ explains $62\%$ of the variation in $\mu$ (Fig.~\ref{fig:theta}, bottom right). The comparison is direct: $\widehat\mu$ is computed without the attention map, and nothing is fitted between the two. The agreement is partial, particularly on object patches, because attention also responds to image-specific key content that the request omits. At the selected head of five of the other seven ViT-B backbones, the requested destination predicts the address as well ($R^2=.57$--$.84$, Appendix~\ref{app:generality_mechanism}).

\begin{table}[t]
\centering
{\footnotesize
\setlength{\tabcolsep}{3.0pt}
\renewcommand{\arraystretch}{1.12}
\begin{tabular}{llrrrr@{\hspace{16pt}}lrr}
  \toprule
  \multicolumn{6}{c}{\emph{Across backbones, held-out COCO}} & \multicolumn{3}{c}{\emph{Different parts of one person}} \\
  \cmidrule(r{10pt}){1-6}\cmidrule(l){7-9}
  & & Original & Request & Requested & Address & & & \\
  Backbone & Head & Q/K & $\theta$ & $\widehat\mu$ & $\mu$ & Readout & Dim. & AUC \\
  \midrule
  DINO & L7h3 & .623 & .644 & \textbf{.661} & .660 & \emph{Vector similarity} & & \\
  DINOv2 & L10h2 & .738 & \textbf{.823} & .821 & .807 & Residual & 768 & .651 \\
  DINOv2-reg & L11h3 & .730 & .805 & \textbf{.806} & .804 & Value & 64 & .720 \\
  DINOv3 & L10h1 & .765 & \textbf{.815} & .814 & .801 & Query & 64 & .856 \\
  MAE & L12h8 & .765 & .830 & \textbf{.831} & .793 & \emph{Where attention goes} & & \\
  CLIP & L9h3 & .746 & \textbf{.769} & .747 & .707 & Attention address $\mu$ & 2 & .935 \\
  SigLIP & L9h2 & .702 & \textbf{.720} & .710 & .627 & Requested destination $\widehat\mu$ & 2 & .940 \\
  Supervised & L11h10 & .609 & \textbf{.614} & .602 & .581 & Full attention pattern & 196 & \textbf{.945} \\
\bottomrule
\end{tabular}}
\caption{\textbf{Where attention goes separates individuals.} \emph{Left:} same-instance AUC at the head selected by the original test in each backbone (Section~\ref{sec:geometrytransfer}). Original Q/K is the routing-side relation selected there, and bold marks the best entry in each row. \emph{Right:} different parts of one person (Section~\ref{sec:addresses}), the two-coordinate address and requested destination nearly match the full attention pattern. Dim.\ is the dimension of the compared vectors.}
\label{tab:transfer}\label{tab:crossparts}
\end{table}

\begin{figure}[t]
  \centering
  \includegraphics[width=0.94\textwidth]{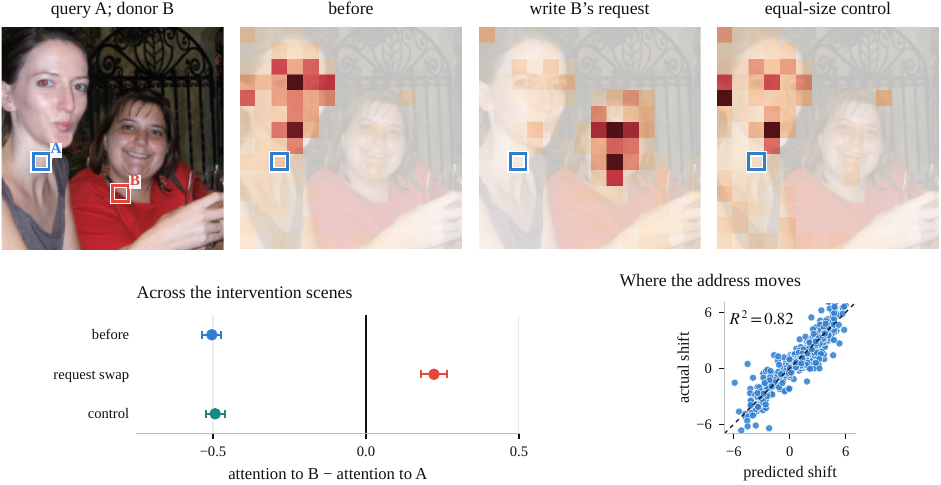}
  \caption{\textbf{Changing the spatial request redirects attention.} Top: a torso patch on person A reads from A. After the request swap, it reads from B, and after the equal-size control, it stays on A. Maps share one color scale, with square-root contrast. Bottom, over all 215 scenes: attention to B minus attention to A (left, from $-1$ for only A to $+1$ for only B) and actual versus predicted address shift (right).}
  \label{fig:validate}
\end{figure}

\section{Editing and grouping with the spatial request}
\label{sec:whatitdoes}

The address can only be observed. The request, in contrast, is read from the native query of the head along five fixed directions, $\theta_i=Bq_i$, so moving the query along them changes the request while the model stays unchanged. We use this to edit the request and to group patches with it.

\subsection{Changing the request redirects attention}
\label{sec:edit}

A request that predicts the address could still be only a correlate of routing. To test whether it controls where the head reads, we took a source patch on person A and a donor patch on person B in each of the 215 COCO scenes, and compared two edits of the source query at L10h2 (Fig.~\ref{fig:validate}). The \emph{request swap} gives the source query the request of the donor, using the smallest query change that does so. The \emph{equal-size control} changes the query by the same amount in random directions that leave its request unchanged. In both cases we edited only this query at this head, kept the image fixed, and reran the model.

After the request swap, attention shifts from A toward B, while after the equal-size control the original route stays largely unchanged. The predicted shift of the requested destination explains $82\%$ of the variation in the actual address shift, again without fitting anything between the two. Changing this component of the native query therefore changes where the head reads, and attention moves gradually as the edit is scaled. The same swap reverses the preference of the selected head in six of the other seven ViT-B backbones and in the larger DINOv2 ViT-L, and in all of these except SigLIP the requested destination also predicts the movement (Appendix~\ref{app:generality_mechanism}). Later instance-sensitive heads respond to this large edit but still prefer person A (Appendix~\ref{app:interventions}), so the edit controls where this head reads, not an identity carried through the network.

\subsection{Grouping repeated objects}
\label{sec:downstream}

If requests carry instance structure, a simple frozen grouping of them should separate repeated objects in a whole image. For each image, we clustered its patches by their request vectors $\theta_i$ with mean-shift, which does not need the number of objects. Its one parameter, a radius, is selected for each representation on labeled VOC development images and then frozen, and no segmentation model is trained (Appendix~\ref{app:downstream}). We score instance assignment only among ground-truth foreground patches, with the foreground adjusted Rand index (FG-ARI).

On all 1,587 eligible COCO scenes, request grouping reaches $.615$ FG-ARI, versus $.289$ for the same clustering of final DINOv2 patch features and $.356$ for the object-centric model SPOT \citep{kakogeorgiou2024spot}. Full attention patterns reach $.649$. Requests therefore better separate repeated instances that feature-based grouping tends to merge, in seven of eight backbones (Appendix~\ref{app:downstream}).


\section{Conclusion}
\label{sec:discussion}

We studied how a pretrained ViT keeps two objects of the same category apart. With category and position controlled, selected heads separate them through where patches read more than through what they carry. Different-looking parts of one person read from a common place, their attention address, while identical copies read from different places. Part of each query, its spatial request, asks for a region of the image. In DINOv2 and most other backbones, this request predicts the address, and changing it redirects the head to the other object. Patches can thus agree on an object by reading from the same place, without looking alike.

\paragraph{Limitations and future work.} Requests separate repeated foreground instances well, but they do not by themselves recover complete object masks: background regions remain fragmented and object boundaries are imperfect. A natural direction is to identify complementary mechanisms for foreground/background separation and object boundaries, and combine them with instance-sensitive routing for stronger object discovery. Another is to test whether pretrained video models use similar routing to maintain object identity over time.

\section*{AI use statement}
In this work, we used generative AI tools to help implement and run experiments and analyze their results, to search for related work, and to draft and polish text. The authors independently checked the reported results and are responsible for the final claims and submission.


\bibliography{references}
\bibliographystyle{iclr2027_conference}

\clearpage
\appendix
\raggedbottom

\section{Data, selection, and the controlled test}
\label{app:protocol}

VOC development images are used to select the components, fit the spatial bases and choose the grouping radii. Disjoint VOC images are used to evaluate those choices. Each COCO test uses every validation image that meets its annotation requirements (Table~\ref{tab:protocolmap}). Parts need visible human keypoints, and grouping needs reliable instance masks.

\begin{center}\begin{minipage}{\textwidth}
\centering\small
\setlength{\tabcolsep}{4pt}\renewcommand{\arraystretch}{1.12}
\begin{tabular}{>{\raggedright\arraybackslash}p{.23\textwidth}>{\raggedright\arraybackslash}p{.31\textwidth}>{\raggedright\arraybackslash}p{.37\textwidth}}
\toprule
Image set & Size and role & Experiments using it \\
\midrule

VOC development & 102 images; 62,362 pairs & Select layers, heads and relations; fit spatial bases without object labels; select grouping radii \\
VOC held-out & 101 images; 71,390 pairs & Native relations, shared reading, spatial requests and controls \\
COCO pair test & 1,414 images; 938,606 pairs & Transfer the same frozen native relations and spatial bases \\
ADE20K pair test & 530 images; 176,732 pairs & Transfer the native routing--value comparison \\
COCO parts & 215 images; 35,100 pairs & Compare different body parts; predict addresses; intervene once per image with five controls \\
COCO grouping & 1,587 images & Group all patches; evaluate instance assignments and object masks \\
Synthetic copies & Three zebra cutouts on three backgrounds & Separate appearance from identity and position \\
DAVIS illustrations & Fixed qualitative examples & Show grouping, retrieval and resolution effects \\
\bottomrule
\end{tabular}
\captionof{table}{\textbf{One controlled question, with specialized tests where needed.} Pair counts include both labels and count each sampled pair once. The COCO tests are overlapping views of one validation population, not independent replications. DAVIS frames illustrate grouping and resolution effects.}
\label{tab:protocolmap}
\end{minipage}\end{center}

\subsection{Which images and patches enter the test?}

We used Pascal VOC 2012 validation images for development and internal evaluation. A fixed 256-image repeated-category list contains 203 images that support the controlled pair test of Section~\ref{sec:controlledtest}. Alternating entries define the 102-image development and 101-image held-out splits, before any component is selected. All spatial fits use the same 102 development images, and all grouping settings are chosen within that split.

An image must contain two instances of one category, each with at least two labeled patches, and at least eight foreground patches overall. A patch receives an instance label when that instance occupies at least $60\%$ of its nonignored pixels. Ambiguous overlapping pixels are ignored, and COCO crowd regions do not count as individual objects. Same-instance pairs are positives, and pairs from different instances of one category are negatives. We matched the horizontal and vertical displacement, including direction: for every included offset, we sampled the same number of positive and negative pairs. We kept at most 12 pairs per label at each offset, so common offsets do not dominate, and required at least 24 pairs per image. These fixed sampling rules determined the counts in Table~\ref{tab:protocolmap}. Every COCO image that meets them is used, and no image is removed because of how a model scores on it. The parts, copy and grouping tests use their own rules (Appendices~\ref{app:crossparts}, \ref{app:copies} and~\ref{app:downstream}).

\subsection{How are matched pairs scored?}

Matching makes each relative offset equally likely under either label, so distance or direction alone cannot solve the test. We evaluated the scores below on these matched pairs.

Each readout gives every pair one score, and a higher score should indicate the same instance. Query, key, value and residual similarities are cosines. Residuals are the patch representations after a transformer block, and their layer is selected separately, by the highest mean same-instance AUC on VOC development. For the query--key relation and direct attention, we averaged both directions because same-instance membership is symmetric:
\begin{equation}
s^{Q\leftrightarrow K}_{ij}=\tfrac12(q_i^\top k_j+q_j^\top k_i),\qquad
s^{A\leftrightarrow}_{ij}=\tfrac12(A_{ij}+A_{ji}).
\label{eq:symmetric}
\end{equation}
Addresses, requested destinations and spatial requests use negative Euclidean distance, and full attention patterns use the overlap of the two attention rows (Appendix~\ref{app:sources}). Each image supplies one AUC, the probability that a random positive pair scores higher than a random negative pair, so no decision threshold is needed. Every image receives equal weight. Confidence intervals use 5,000 resamples of images, not of individual patch pairs.

\subsection{Which native relations carry the advantage?}

Table~\ref{tab:operands} expands the comparison of Figure~\ref{fig:qkv}. For each relation, a layer and head were selected on VOC development images and then used unchanged on both held-out sets. No choice uses test outcomes.

\begin{center}\begin{minipage}{\textwidth}
\centering\small\setlength{\tabcolsep}{6pt}
\begin{tabular}{llrrrrrr}
\toprule
Set & Backbone & Residual & Query & Key & Query--key & Attention & Value \\
\midrule
VOC & DINO & .593 & \textbf{.644} & .620 & .642 & .627 & .598 \\
 & DINOv2 & .563 & .740 & .734 & .741 & \textbf{.795} & .636 \\
 & DINOv2-reg & .584 & .749 & .770 & .786 & \textbf{.791} & .605 \\
 & DINOv3 & .605 & .801 & .747 & .782 & \textbf{.810} & .596 \\
 & MAE & .621 & .777 & .764 & .794 & \textbf{.812} & .700 \\
 & CLIP & .627 & .715 & .722 & .755 & \textbf{.779} & .648 \\
 & SigLIP & .550 & .655 & .650 & .727 & \textbf{.737} & .581 \\
 & Supervised & .596 & .623 & .605 & .613 & \textbf{.648} & .598 \\
\midrule
COCO & DINO & .586 & \textbf{.623} & .606 & .623 & .608 & .588 \\
 & DINOv2 & .565 & .707 & .693 & .690 & \textbf{.738} & .616 \\
 & DINOv2-reg & .565 & .700 & .722 & \textbf{.732} & .730 & .589 \\
 & DINOv3 & .602 & \textbf{.765} & .725 & .746 & .765 & .606 \\
 & MAE & .610 & .725 & .699 & .742 & \textbf{.765} & .659 \\
 & CLIP & .611 & .703 & .700 & .728 & \textbf{.746} & .631 \\
 & SigLIP & .562 & .625 & .620 & .693 & \textbf{.702} & .584 \\
 & Supervised & .594 & .609 & .612 & .609 & \textbf{.625} & .592 \\
\bottomrule
\end{tabular}
\captionof{table}{\textbf{The routing advantage appears in several native relations.} Mean same-instance AUC on 101 VOC and 1,414 COCO images. Bold marks the strongest routing relation within each row. All choices use the development set.}
\label{tab:operands}
\end{minipage}\end{center}

The right panel of Figure~\ref{fig:qkv} asks a complementary question: at the same instance-sensitive heads, what do the routing side and the values preserve best? It compares three pair tests on the 101 held-out VOC images, each with displacement matched. The instance test is the one of Section~\ref{sec:controlledtest}. In the foreground test, a positive pair contains two foreground patches and a negative pair contains one foreground and one background patch. In the category test, both patches of every pair come from different instances, and only positive pairs share a category. All 101 images support the foreground test and 47 support the category test. Background here means no reliable foreground assignment, which includes mixed boundary patches, so this row compares patch pairs and is not a foreground segmentation.

\subsection{Does the result extend beyond the small VOC split?}

The development split is small on purpose, 102 VOC images. If instance-sensitive routing is already part of how a pretrained ViT works, a few images should be enough to find the heads that carry it, and those heads should then work unchanged on larger datasets. They do. With every choice frozen on VOC, routing exceeds value similarity in all eight backbones on the 1,414-image COCO test (Figure~\ref{tab:robust}). On the 530-image ADE20K test, routing also exceeds value similarity in every backbone.

\begin{center}\begin{minipage}{\textwidth}
\centering
\includegraphics[width=.68\textwidth]{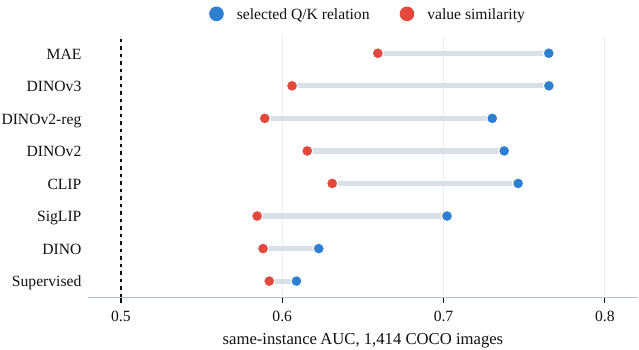}
\captionof{figure}{\textbf{Routing remains stronger than value similarity on the larger COCO test.} Blue and red show the actual AUCs of the independently VOC-selected routing and value components.}
\label{tab:robust}
\end{minipage}\end{center}

\subsection{Model inputs and reproducibility}
\label{app:reproduction}

All backbones are frozen public \emph{timm} (Python Image Models) checkpoints in evaluation mode (Table~\ref{tab:checkpoints}). Layers are numbered from 1 and heads from 0. Inputs are resized on the shorter side and center-cropped to a square: $196$ pixels for patch-14 models and $224$ for patch-16 models, giving $14\times14$ patches in both cases. Image resizing is bilinear and masks use nearest-neighbor resizing. Inputs use the channel normalization of each checkpoint. Learned absolute position embeddings are interpolated by \emph{timm} when the grid changes, and DINOv3 uses its native rotary positions. DAVIS illustrations resize the full frame to a square. The higher-resolution examples use $392$-pixel DINOv2 inputs and a $28\times28$ grid with the same basis and radii.

Q, K and V are captured at scaled dot-product attention, after the Q/K normalization of the architecture and the DINOv3 rotary embeddings. Prefix and register tokens take part in native attention. Spatial distributions then discard those destinations and renormalize over image patches, both for the attention address and for the requested destination.

\begin{center}\begin{minipage}{\textwidth}
\centering\small
\setlength{\tabcolsep}{7pt}
\begin{tabular}{ll}
\toprule
Model & Checkpoint \\
\midrule
DINO & vit\_base\_patch16\_224.dino \\
DINOv2 & vit\_base\_patch14\_dinov2.lvd142m \\
DINOv2-reg & vit\_base\_patch14\_reg4\_dinov2.lvd142m \\
DINOv3 & vit\_base\_patch16\_dinov3.lvd1689m \\
MAE & vit\_base\_patch16\_224.mae \\
OpenCLIP & vit\_base\_patch16\_clip\_224.openai \\
SigLIP & vit\_base\_patch16\_siglip\_224.webli \\
AugReg & vit\_base\_patch16\_224.augreg\_in21k\_ft\_in1k \\
\bottomrule
\end{tabular}
\captionof{table}{Frozen backbones used in the eight-model comparison.}
\label{tab:checkpoints}
\end{minipage}\end{center}

\newpage
\section{Where does instance-sensitive routing appear?}
\label{app:qk}

The main comparison selects components on development images. Figure~\ref{fig:allheadmaps} shows the surrounding heads on the held-out set: strong instance discrimination is concentrated in a few heads rather than spread across attention. Figure~\ref{fig:layers} shows how the address field changes with depth in one fixed scene. The most instance-discriminative head can change with the scoring convention, so localization does not imply a unique head.

\begin{center}\begin{minipage}{\textwidth}
\centering
\includegraphics[width=.95\textwidth]{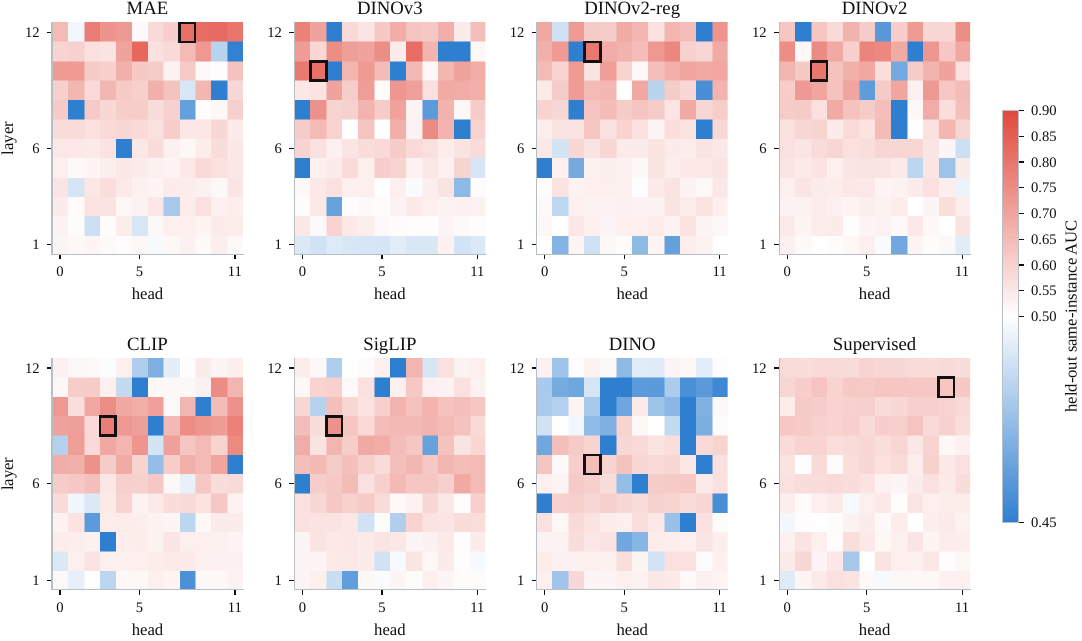}
\captionof{figure}{\textbf{Instance-sensitive routing is localized.} Each cell is the held-out same-instance AUC of one layer and head. The outlined square marks the component selected on the development split. Strong heads appear at different depths across pretraining methods, and most heads have weaker instance discrimination.}
\label{fig:allheadmaps}
\end{minipage}\end{center}
\begin{center}\begin{minipage}{\textwidth}
\centering
\includegraphics[width=.92\textwidth]{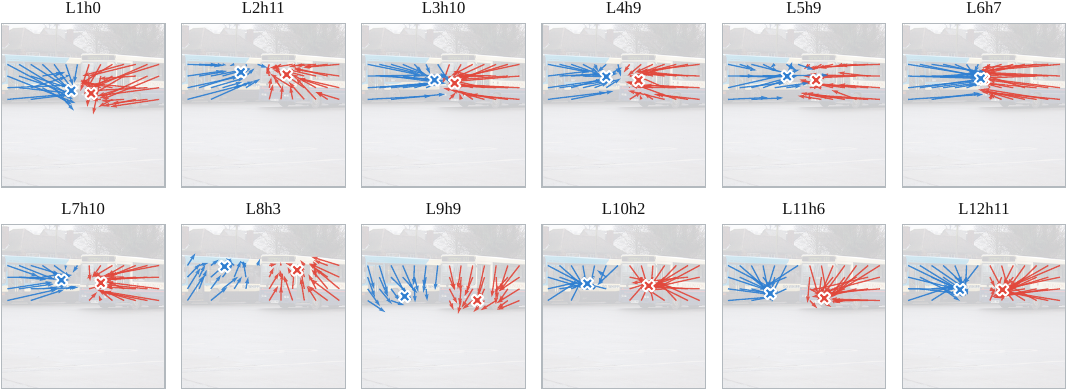}
\captionof{figure}{\textbf{Object-specific address fields emerge through depth.} At each layer the head is selected by direct-attention AUC on the same development split. All panels show the same held-out scene. Later DINOv2 heads form two compact destinations, but the pattern is neither universal nor monotonic.}
\label{fig:layers}
\end{minipage}\end{center}

\newpage
\section{What does the attention address capture?}
\label{app:sources}

This appendix supports Section~\ref{sec:addresses}. It compares full attention patterns with direct attention, tests whether locality alone explains the addresses, and describes the two tests of what the address tracks: different parts of one person and identical copies.

\subsection{Is shared reading more informative than direct attention?}

We compared the full attention patterns of two patches with the Bhattacharyya coefficient, $\sum_k\sqrt{\widetilde A_{ik}\widetilde A_{jk}}$ \citep{bhattacharyya1946divergence}. This classical overlap measure has also been used to compare attention vectors across BERT heads \citep{bian2021redundancy}. Here it compares different patches within one head. Table~\ref{tab:overlap} keeps the head selected by the original instance test fixed. Full attention patterns (shared reading) separate instances better than direct attention in 14 of 16 model--dataset comparisons.

\begin{center}\begin{minipage}{\textwidth}
\centering\small
\setlength{\tabcolsep}{4.5pt}
\begin{tabular}{lrrrrrr}
\toprule
& \multicolumn{3}{c}{Held-out VOC} & \multicolumn{3}{c}{COCO} \\
\cmidrule(lr){2-4}\cmidrule(lr){5-7}
& Full attention & Direct & & Full attention & Direct & \\
Backbone & pattern & attention & Gain & pattern & attention & Gain \\
\midrule
DINO & .667 & .627 & $+.041$ & .631 & .608 & $+.023$ \\
DINOv2 & .873 & .795 & $+.078$ & .818 & .738 & $+.081$ \\
DINOv2-reg & .853 & .791 & $+.063$ & .792 & .730 & $+.062$ \\
DINOv3 & .852 & .810 & $+.042$ & .812 & .765 & $+.047$ \\
MAE & .833 & .812 & $+.021$ & .770 & .765 & $+.005$ \\
CLIP & .789 & .779 & $+.010$ & .756 & .746 & $+.009$ \\
SigLIP & .705 & .737 & $-.032$ & .675 & .702 & $-.027$ \\
Supervised & .636 & .495 & $+.141$ & .610 & .514 & $+.096$ \\
\bottomrule
\end{tabular}
\captionof{table}{\textbf{Full attention patterns strengthen the same-instance relation.} Same-instance AUC at the head selected by the original test in each backbone, on the same matched pairs.}
\label{tab:overlap}
\end{minipage}\end{center}

\subsection{Could local attention explain the addresses?}

Could nearby patches share an address simply because attention stays nearby? We built a distance-only attention map for each query: nearby destinations receive more weight than distant ones, regardless of the image. We set its Gaussian width so that attention travels the same average distance as in the real head. A second control uses the attention maps that the head produces for another image, placed on the current grid. This keeps realistic reading patterns while breaking their connection to the current objects.

Both controls use the original 101 held-out images, selected heads and matched patch pairs. For reproducibility, each image borrows attention from the previous image in the saved list, with the first borrowing from the last. We compared patches by the distance between their resulting addresses. In DINOv2, real addresses reach $.870$ AUC, versus $.518$ for distance-only attention and $.523$ for attention from another image. The real addresses outperform both controls in all eight backbones (Table~\ref{tab:locality}). They therefore reflect the current scene, beyond nearby patches tending to read nearby. Figures~\ref{fig:address_examples} and~\ref{fig:address_examples_dinov3} show visual examples.

\begin{center}\begin{minipage}{\textwidth}
\centering\small\setlength{\tabcolsep}{4.2pt}
\begin{tabular}{lcccccccc}
\toprule
Address from & DINO & DINOv2 & DINOv2-reg & DINOv3 & MAE & CLIP & SigLIP & Supervised \\
\midrule
Real attention & \textbf{.708} & \textbf{.870} & \textbf{.859} & \textbf{.847} & \textbf{.876} & \textbf{.744} & \textbf{.636} & \textbf{.616} \\
Distance only & .523 & .518 & .520 & .518 & .514 & .519 & .507 & .509 \\
Another image & .520 & .523 & .516 & .529 & .505 & .524 & .508 & .499 \\
\bottomrule
\end{tabular}
\captionof{table}{\textbf{Locality does not explain the addresses.} Same-instance AUC of the attention address on the 101 held-out VOC images, at the head selected in each backbone.}
\label{tab:locality}
\end{minipage}\end{center}

\subsection{Different parts of the same person}
\label{app:crossparts}

Can two visually different parts of the same person share a destination? COCO keypoints let us test this while keeping the original same-instance comparison. We assigned patches to head, torso, arm or leg using visible keypoints: head and face keypoints define the head, shoulders and hips the torso, elbows and wrists the arms, and knees and ankles the legs. To avoid ambiguous assignments, patches must have at least $90\%$ instance purity, contain no annotated overlap, and lie within 2.5 patches of the nearest visible keypoint. These rules depend only on annotations, not on model scores. The labels approximate parts from keypoints and are not dense part masks.

Both positive and negative pairs connect different body parts. Within each image, we balanced the labels at every ordered part pair and exact signed offset, keeping up to 24 of each label per group, and required at least 12 pairs. Scanning all COCO validation images yields 215 scenes and 35,100 pairs. The same scenes support the intervention of Appendix~\ref{app:interventions}. On these pairs the request nearly matches the full attention pattern ($.942$ versus $.945$ AUC).

\subsection{Identical copies}
\label{app:copies}

The copy test of Figure~\ref{fig:copies} removes appearance as an explanation for different addresses. Three complete COCO zebra cutouts were each placed on three natural backgrounds, giving nine combinations. Within a scene the two copies are pixel-identical, and only patches covered by the object mask are scored. All nine combinations yield separate, correctly ordered, movable addresses in DINOv2 (L10h2) and DINOv3 (L10h1). When one copy moves, its address shifts by $4.68$ and $4.61$ patches on average, respectively, while the address of the stationary copy shifts by $.98$ and $1.22$. Figure~\ref{fig:copies} uses a fourth background, licensed for reproduction, and the zebra with the median score of the three.

\section{How is the spatial request constructed and validated?}
\label{app:theta}

\subsection{Fitting the spatial basis}

We fitted the average key at each patch position using six maps, $\phi=(1,r,c,r^2,c^2,rc)$. On the $14\times14$ grid, $r$ and $c$ are the row and column coordinates, each centered and divided by its standard deviation. Keys are averaged at each position over the 102 development images. We find one key vector $b_t$ for each map $\phi_t$ so that their weighted sum reconstructs these average keys:
\begin{equation}
\{b_t\}_{t=0}^{5}=\arg\min_{\{u_t\}}\sum_j\left\|\bar k(r_j,c_j)-\sum_{t=0}^{5}\phi_t(r_j,c_j)u_t\right\|_2^2,
\label{eq:lstsq}
\end{equation}
where $j$ indexes patches and each $u_t$ has 64 entries. The constant vector $b_0$ cancels in the softmax. The other five vectors form the rows of the $5\times64$ matrix $B$ of Section~\ref{sec:construction}, so the five entries of $\theta_i=Bq_i$ are the dot products of the query with these vectors. We froze the fitted vectors and reused them for evaluation and illustrations. Labels selected the head, but the fit itself uses only coordinates and native keys. Figure~\ref{fig:theta} shows how a query weights the five maps to produce its request.

\subsection{Why use a quadratic function if two coefficients already separate instances?}

Two slopes already separate instances well. Why keep curvature? Slopes can only tilt attention toward one side of the image, while curvature lets a query place a peak of attention around one region, such as its own object (Section~\ref{sec:construction}). It also helps explain where attention goes, how it moves under edits, and which patches group together. Table~\ref{tab:curvature} compares linear, quadratic and cubic descriptions of the same head.

\begin{center}\begin{minipage}{\textwidth}
\centering\small\setlength{\tabcolsep}{3pt}
\begin{tabular}{lrrrrrrr}
\toprule
& Key fit & \multicolumn{2}{c}{Same-instance AUC $\uparrow$} & Address & Movement & \multicolumn{2}{c}{Grouping $\uparrow$} \\
\cmidrule(lr){3-4}\cmidrule(lr){7-8}
Spatial description & $R^2$ $\uparrow$ & VOC & Person parts & error $\downarrow$ & $R^2$ $\uparrow$ & FG-ARI & mIoU \\
\midrule
Linear & .669 & .871 & .909 & 2.11 & .646 & .556 & .301 \\
Quadratic & .907 & \textbf{.872} & \textbf{.940} & \textbf{1.90} & .821 & \textbf{.615} & \textbf{.429} \\
Cubic & \textbf{.941} & .869 & .939 & 1.92 & \textbf{.827} & .594 & .406 \\
\bottomrule
\end{tabular}
\captionof{table}{\textbf{Curvature helps explain routing and grouping.} Key fit is the share of the variation in the development mean keys that each description explains. Same-instance AUC compares requests on held-out VOC and requested destinations on different parts of one person (Appendix~\ref{app:crossparts}). Address error, in patches, and movement $R^2$ use the 215 part scenes and the request swaps of Appendix~\ref{app:interventions}, with nothing fitted to the outcomes. Grouping uses the rule of Appendix~\ref{app:downstream} on all 1,587 COCO images. Bold marks the best value in each column.}
\label{tab:curvature}
\end{minipage}\end{center}

Curvature barely changes pairwise discrimination on VOC, but it improves the address prediction, the movement prediction and grouping. Cubic terms fit the average keys better still, but bring little change in routing prediction and no gain in grouping. These results favor the quadratic as a compact description. 

\subsection{Does the request retain the instance relation?}

Table~\ref{tab:strictgeometry} supports Section~\ref{sec:geometrytransfer}. At the DINOv2 L10h2 head, the spatial request and the requested destination retain the same-instance relation on both held-out sets.

\begin{center}\begin{minipage}{\textwidth}
\centering
\begin{minipage}[c]{0.42\textwidth}
\centering\footnotesize
\setlength{\tabcolsep}{5pt}
\begin{tabular}{lcc}
\toprule
& VOC & COCO \\
\midrule
Residual similarity & .563 & .565 \\
Query similarity & .740 & .707 \\
Direct attention & .795 & .738 \\
Attention address $\mu$ & .870 & .807 \\
Spatial request $\theta$ & .872 & .823 \\
Requested destination $\widehat\mu$ & .872 & .821 \\
Full attention pattern & .873 & .818 \\
Slopes only & .871 & .805 \\
Curvature only & .628 & .655 \\
\bottomrule
\end{tabular}
\end{minipage}\hspace{\fill}%
\begin{minipage}[c]{0.55\textwidth}
\centering
\includegraphics[width=\linewidth]{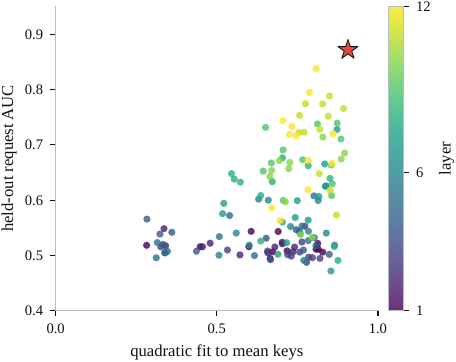}
\end{minipage}
\par\vspace{3pt}
\begin{minipage}[t]{0.42\textwidth}
\captionof{table}{\textbf{The spatial request retains the relation that selected the head.} Same-instance AUC on held-out VOC and COCO.}
\label{tab:strictgeometry}
\end{minipage}\hspace{\fill}%
\begin{minipage}[t]{0.55\textwidth}
\captionof{figure}{\textbf{Keys fitted with quadratic alone are not enough.} Each point is one of the 144 heads of DINOv2, with its own development fit and request AUC on held-out VOC. The selected head (star) has the strongest fit and request AUC, while many heads with quadratic keys carry little instance signal.}
\label{fig:lockedgeometry}
\end{minipage}
\end{minipage}\end{center}

Three additional controls examine simpler explanations. \emph{Architecture alone:} the same construction in a seeded untrained network gives request AUCs of $.517$ on VOC and $.537$ on COCO. \emph{A fixed spatial pattern:} assigning requests from another image to the current grid gives $.530$ and $.513$. \emph{Category imbalance:} also matching the actual category within each offset leaves $.881$ and $.841$, on the 95 VOC and 1,263 COCO images with enough balanced pairs. The signal therefore depends on learned features of the current scene and survives matching category as well as relative position.

Figure~\ref{fig:lockedgeometry} details whether keys fitted with quadratic alone are sufficient. They are not: many heads have a good quadratic fit but little instance discrimination. Fit and request AUC correlate at $.38$ across all 144 heads but only $.06$ within layers, so depth explains much of their association.

\noindent\begin{minipage}{\textwidth}
\subsection{How close is the requested destination to the attention address?}

Table~\ref{tab:predictioncontrols} compares the requested destination with the actual attention address, as the mean distance in patches over all 1,414 COCO images. Two controls are used to give reference errors. \emph{Wrong image} borrows the requested destination at the same grid position from another image. \emph{Source position} predicts that each patch reads at its own location. For every backbone, the current request predicts better than both controls, although the remaining error varies.

\begin{center}\begin{minipage}{\textwidth}
\centering\small
\setlength{\tabcolsep}{5pt}
\begin{tabular}{lrrr}
\toprule
Backbone & Current request & Wrong image & Source position \\
\midrule
DINO & 1.32 & 1.95 & 2.88 \\
DINOv2 & 1.74 & 2.59 & 3.36 \\
DINOv2-reg & 1.55 & 2.58 & 3.38 \\
DINOv3 & 1.42 & 2.77 & 3.34 \\
MAE & 1.17 & 2.04 & 3.79 \\
CLIP & 1.79 & 2.65 & 3.10 \\
SigLIP & 2.76 & 3.27 & 4.57 \\
Supervised & 2.38 & 2.54 & 5.11 \\
\bottomrule
\end{tabular}

\captionof{table}{\textbf{The current request predicts attention destinations better than spatial controls.} Mean Euclidean error, in patches, between each predicted destination and the attention address over all patches of 1,414 COCO images.}
\label{tab:predictioncontrols}
\end{minipage}\end{center}
\end{minipage}

\subsection{Why does the request not reproduce all of attention?}

The spatial basis describes the average key at each position well, but the keys of one image also carry what that image contains. Let $e_j$ be the part of the actual key left out by the fitted basis, and let $\phi_j=(r_j,c_j,r_j^2,c_j^2,r_jc_j)^\top$ contain the five position-map values. Then
\begin{align}
k_j &= b_0+B^\top\phi_j+e_j,\\
q_i^\top k_j &= q_i^\top b_0+\underbrace{\theta_i^\top\phi_j}_{F_i(r_j,c_j)}+q_i^\top e_j.
\label{eq:remainder}
\end{align}
The first term is constant over destinations and cancels in the softmax. The middle term produces the request distribution $\widehat A_i$. The last term reweights it according to the key structure that the fit omits:
\begin{equation}
\widetilde A_{ij}=\frac{\widehat A_{ij}\exp(q_i^\top e_j/\sqrt d)}
{\sum_{t\in\mathcal P}\widehat A_{it}\exp(q_i^\top e_t/\sqrt d)}.
\end{equation}
Here $\mathcal P$ is the set of image patches and $d=64$ for this head. The remainder includes image-specific key variation and errors in the mean-key fit. Changing a query can change both its request and its interaction with the remainder. The request swap of Section~\ref{sec:edit} therefore establishes control by that query component. It does not isolate the quadratic score as the only cause of the attention shift.

\newpage
\section{How is the request edited, and does the mechanism hold beyond DINOv2?}
\label{app:interventions}

\subsection{Changing one query}

We wanted to change the part of a query that sets its spatial request. To do so, we took the request of a donor patch on person B and wrote it into the query of a source patch on person A, using the smallest query change that achieved this. Every one of the 215 part scenes contributed its first sampled different-instance pair as source and donor, and no scene or pair is selected for successful redirection. If $B^+$ is the pseudoinverse of $B$, the request swap is
\begin{equation}
\Delta q=B^+(\theta_B-\theta_A).
\end{equation}
This sets the request of the source to $\theta_B$ with the smallest change to its query, on average $.60$ of the query norm. The equal-size controls are five random query changes of the same length that satisfied $B\Delta q=0$, so the request stays unchanged, and we averaged their results within each image. Figure~\ref{fig:validate} uses a torso pair for illustration, and its aggregate plots use all 215 predetermined pairs. Figure~\ref{fig:swapattention} shows one edit in more detail.

\begin{center}\begin{minipage}{\textwidth}
\centering
\includegraphics[width=.9\textwidth]{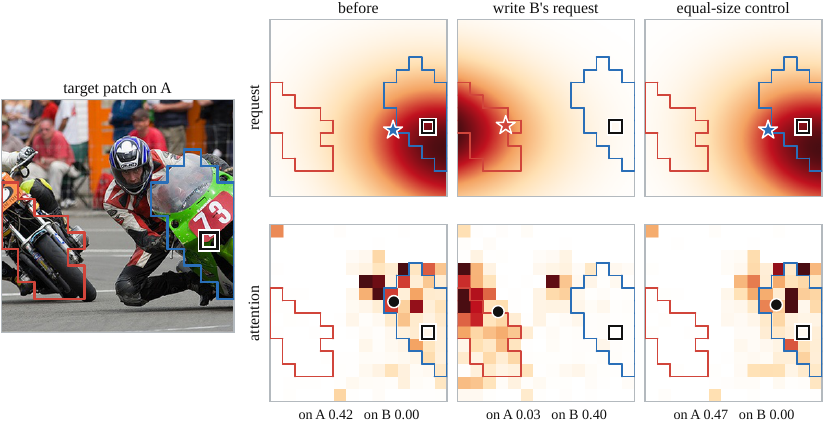}
\captionof{figure}{\textbf{The edited request moves where the patch reads.} Top: the request of one patch on rider A before the edit, after writing the mean request of rider B, and after an equal-size control edit that leaves the request unchanged. Bottom: the attention of that patch in the same three conditions, with its attention address ($\bullet$) and the attention mass on each rider.}
\label{fig:swapattention}
\end{minipage}\end{center}

The requested destination predicts the direction of the address movement in 213 of 215 scenes. The address moves $3.68$ patches on average, against $3.81$ predicted and $.40$ under the equal-size controls. We measure the effect with B$-$A, the attention that the source patch places on the patches of person B minus the attention it places on person A, with attention renormalized over image patches. It ranges from $-1$ (the patch reads only from A) to $+1$ (only from B), so negative values mean more attention on A than on B, and positive values the reverse. The effect grows smoothly with the size of the edit: scaling $\Delta q$ by $.25$, $.5$, $.75$, $1$ and $1.5$ moves B$-$A from $-.50$ to $-.40$, $-.21$, $+.03$, $+.22$ and $+.42$ (Figure~\ref{fig:validate} shows the full-size swap). Writing instead the request of another patch of person A, the one most different from the source, leaves attention on A ($-.41$). Random changes of the same size inside the space of requests move the address less predictably than the request of a real donor, so the prediction concerns the requests that patches actually make.

\begin{center}\begin{minipage}{\textwidth}
\centering\small
\setlength{\tabcolsep}{7pt}
\begin{tabular}{lccc}
\toprule
Condition & Edited L10h2 & Selected L11h6 & Selected L12h11 \\
\midrule
Intact & $-0.504$ & $-0.361$ & $-0.280$ \\
$\theta$ swap & $+0.222$ & $-0.251$ & $-0.242$ \\
Matched controls & $-0.493$ & $-0.360$ & $-0.280$ \\
\bottomrule
\end{tabular}
\captionof{table}{\textbf{Editing the query component redirects this head.} Attention toward person B minus person A, averaged over all 215 part scenes. Later heads are selected for instance sensitivity on VOC development. Their attention shifts toward B but still favors A.}
\label{tab:causal}
\end{minipage}\end{center}

The later heads of Table~\ref{tab:causal} respond to the edit but keep their original preference. The result supports control of the reading of the edited head, not a lasting reassignment of identity.

\subsection{Does the mechanism hold in other backbones?}
\label{app:generality_mechanism}

We repeated the analysis at the development-selected head of every backbone, with its own frozen spatial basis, on the same 215 part scenes, pairs, donors and control seeds. We also ran DINOv2 ViT-L/14, whose head is selected by the same rule (L22h11). In most backbones, the DINOv2 pattern repeats (Table~\ref{tab:mechanism_all}). The request separates different parts of one person better than query, value and final-feature similarity, and its destination predicts the address. The request swap reverses the preference of the head while equal-size controls do not, and the requested destination predicts how far the address moves. Two other DINOv3 heads that the rule selects in some re-splits, L11h7 and L12h0, give the same result.

Two backbones differ. In SigLIP, the swap still reverses the preference, but the movement is poorly predicted. Its head sends about a third of the attention of each patch to one single patch, an attention sink \citep{darcet2024registers}, and SigLIP has no CLS or register token to absorb it. Once the two most-attended patches of each image are left out, the address and the movement are predicted again ($R^2$ of $.60$ and $.65$). In the supervised model, the selected head L11h10 shifts less and keeps its preference. This head is atypical: the same rule picks it in only $8\%$ of random re-splits of the development images, and picks L9h8 in most of the others. At L9h8 the swap reverses the preference and the movement is predicted ($R^2=.67$).

\begin{center}\begin{minipage}{\textwidth}
\centering\small\setlength{\tabcolsep}{2.5pt}
\resizebox{\textwidth}{!}{\begin{tabular}{llcccccccc}
\toprule
& & \multicolumn{2}{c}{Person parts AUC} & Address & \multicolumn{2}{c}{B $-$ A} & Movement & \multicolumn{2}{c}{Grouping FG-ARI} \\
\cmidrule(lr){3-4}\cmidrule(lr){6-7}\cmidrule(lr){9-10}
Backbone & Head & Request & Best other & $R^2$ & Swap & Control & $R^2$ & Request & Features \\
\midrule
DINO & L7h3 & .634 & .625 & .84 & $-.24\to+.09$ & $-.23$ & .84 & .398 & .194 \\
DINOv2 & L10h2 & .942 & .856 & .62 & $-.50\to+.22$ & $-.49$ & .82 & .615 & .289 \\
DINOv2-reg & L11h3 & .949 & .883 & .72 & $-.50\to+.36$ & $-.49$ & .89 & .585 & .265 \\
DINOv3 & L10h1 & .944 & .860 & .79 & $-.48\to+.15$ & $-.47$ & .88 & .585 & .380 \\
MAE & L12h8 & .942 & .762 & .76 & $-.39\to+.16$ & $-.38$ & .89 & .617 & .177 \\
CLIP & L9h3 & .895 & .838 & .57 & $-.54\to+.31$ & $-.53$ & .71 & .511 & .209 \\
SigLIP & L9h2 & .840 & .734 & $-.16$ & $-.20\to+.13$ & $-.20$ & $-.21$ & .445 & .154 \\
Supervised & L11h10 & .658 & .699 & .04 & $-.15\to-.06$ & $-.15$ & $-.06$ & .233 & .252 \\
Supervised$^\dagger$ & L9h8 & .593 & .680 & .52 & $-.37\to+.10$ & $-.36$ & .67 & .273 & .252 \\
\midrule
DINOv2 ViT-L & L22h11 & .955 & .842 & .35 & $-.32\to+.15$ & $-.31$ & .74 & --- & --- \\
\bottomrule
\end{tabular}}
\captionof{table}{\textbf{The request mechanism across backbones.} All columns except grouping use the 215 COCO part scenes. Request is the spatial request $\theta$. Best other is the strongest of query, value and final-feature similarity. B $-$ A is attention to person B minus attention to person A, before and after the request swap (Swap) and after the equal-size control (Control). Address $R^2$ compares requested and actual destinations, and movement $R^2$ compares predicted and actual address shifts under the request swap, with nothing fitted. Grouping uses the rule of Appendix~\ref{app:downstream} on all 1,587 COCO grouping images. $^\dagger$The head chosen most often across 1,000 random development re-splits.}
\label{tab:mechanism_all}
\end{minipage}\end{center}

\section{What does request grouping recover?}
\label{app:downstream}

\subsection{One grouping rule for every representation}

Every representation was grouped by the same rule, mean-shift clustering. Mean-shift repeatedly moves each patch vector to the average of the vectors within a fixed radius, and vectors that settle at the same point form one group. The number of groups is not given in advance: it follows from the radius and the image. The vectors of each image were rescaled so that one radius can serve all images. For each representation, we chose the radius on the 96 grouping-eligible VOC development images, as the one of eleven radii between $.15$ and $1$ with the highest mean FG-ARI plus mIoU (Figure~\ref{fig:radius}). The chosen radii then stay fixed on every COCO image.

\begin{center}\begin{minipage}{\textwidth}
\centering
\includegraphics[width=.8\textwidth]{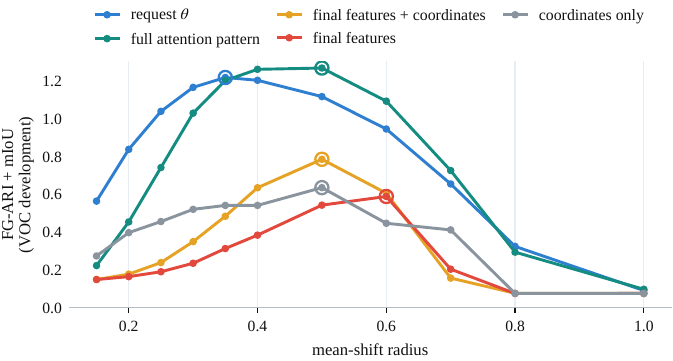}
\captionof{figure}{\textbf{Choosing the radius.} Mean FG-ARI plus mIoU on the 96 VOC development images for each mean-shift radius, DINOv2. Circles mark the radius kept for each representation.}
\label{fig:radius}
\end{minipage}\end{center}

We grouped the spatial request $\theta_i$ and the full attention pattern of the selected head, and the final patch features. Two further baselines verify whether position alone explains request grouping: patch coordinates alone, and final features with patch coordinates appended, whose weight was chosen with the radius. We also evaluated the released COCO model of SPOT \citep{kakogeorgiou2024spot}, trained with seven slots on DINO ViT-B/16 features, by assigning each patch to its strongest slot. Evaluated with its own code on all 5,000 COCO validation images, this checkpoint reproduces its published scores (mBO$_i$ $.349$ and mBO$_c$ $.448$).

\subsection{Instance assignments and object masks}

All 1,587 grouping images contain two same-category instances with at least four reliable patches each. The $14\times14$ grid uses unique instance IDs and $60\%$ patch purity. Crowd-only pixels are ignored and smaller individual masks take precedence in overlaps. The pair test instead excludes ambiguous overlap pixels, so the two tests can select different images. Every method assigns every patch before scoring. FG-ARI measures instance agreement among foreground patches. mIoU scores one-to-one object-mask matches, and mBO$_i$ lets each object choose its best mask. Both mask scores also penalize groups that spill into the background or leave an object incomplete.

\begin{center}\begin{minipage}{\textwidth}
\centering\small\setlength{\tabcolsep}{6pt}
\begin{tabular}{lrrr}
\toprule
Grouping rule & FG-ARI & mIoU & mBO$_i$ \\
\midrule
Spatial request $\theta$ & .615 & \textbf{.429} & \textbf{.451} \\
Full attention pattern & \textbf{.649} & .403 & .436 \\
Final features & .289 & .317 & .356 \\
Final features + coordinates & .403 & .375 & .403 \\
Patch coordinates only & .325 & .229 & .251 \\
SPOT (trained, seven slots) & .356 & .288 & .350 \\
\bottomrule
\end{tabular}
\captionof{table}{\textbf{Routing separates repeated instances better than final features and SPOT, but its masks remain incomplete.} All 1,587 eligible COCO images. The first five rows use the same DINOv2 backbone and mean shift with radii chosen on VOC development. SPOT is the released COCO model, trained on DINO features with seven slots.}
\label{tab:coco}
\end{minipage}\end{center}

Request grouping separates repeated instances far better than final features ($.615$ versus $.289$ FG-ARI, higher in $84\%$ of images). Position alone does not explain this: patch coordinates reach $.325$, and final features with coordinates added reach $.403$. SPOT, which learns object slots on COCO, reaches $.356$, with lower mask scores than the request. Scored instead with the evaluation code of SPOT, the ordering on these images is the same (Table~\ref{tab:spotofficial}). On all 5,000 COCO validation images, SPOT gives better masks, while requests still separate instances better. On the images with repeated objects, SPOT keeps a high class-level score (mBO$_c$) but a low instance-level score (mBO$_i$), as when one slot covers several objects of one category. On DINO, the backbone that SPOT itself uses, request grouping reaches $.398$ (Table~\ref{tab:mechanism_all}). The full attention pattern keeps instances apart slightly better ($.649$), while requests give better object masks. Mask scores stay modest for every representation without training, because background patches also form groups.  Figure~\ref{fig:grouping} shows examples with and without the background. Across backbones, request grouping exceeds final-feature grouping in seven of eight (Table~\ref{tab:mechanism_all}). The exception is the selected head of the supervised model. \begin{center}\begin{minipage}{\textwidth} \centering\small\setlength{\tabcolsep}{4.5pt} \begin{tabular}{lrrrrrrrr} \toprule & \multicolumn{4}{c}{1,587 images with repeated objects} & \multicolumn{4}{c}{All 5,000 COCO validation images} \\ \cmidrule(lr){2-5}\cmidrule(lr){6-9} Grouping rule & FG-ARI & mIoU & mBO$_i$ & mBO$_c$ & FG-ARI & mIoU & mBO$_i$ & mBO$_c$ \\ \midrule Spatial request $\theta$ & \textbf{.548} & \textbf{.307} & \textbf{.322} & .302 & \textbf{.410} & .291 & .300 & .319 \\ Final features & .264 & .227 & .253 & .381 & .264 & .282 & .295 & .374 \\ Final features + coordinates & .364 & .264 & .282 & .308 & .309 & .267 & .278 & .317 \\ SPOT & .336 & .222 & .266 & \textbf{.441} & .368 & \textbf{.328} & \textbf{.349} & \textbf{.448} \\ \bottomrule \end{tabular} \captionof{table}{\textbf{The comparison with SPOT under its own evaluation code.} Masks at $320$ pixels with overlapping pixels ignored, as in SPOT, and our $14\times14$ groups upsampled to $320$ pixels. On repeated objects, requests separate instances better. On general COCO images, SPOT gives better masks.} \label{tab:spotofficial} \end{minipage}\end{center}

\begin{center}\begin{minipage}{\textwidth}
\centering
\includegraphics[width=.6\textwidth]{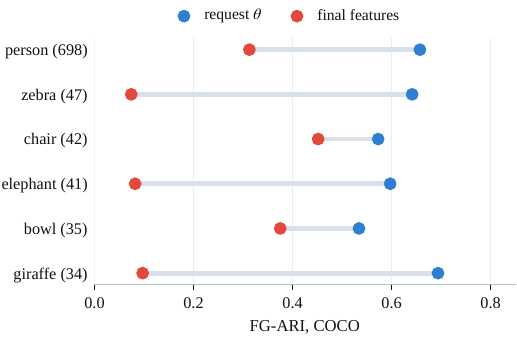}
\captionof{figure}{\textbf{Request grouping keeps its advantage in common categories.} FG-ARI for the six most frequent categories of the COCO grouping set (number of images in parentheses), DINOv2 with the rule of Table~\ref{tab:coco}.}
\label{fig:categories}
\end{minipage}\end{center}

\begin{center}\begin{minipage}{\textwidth}
  \centering
  \includegraphics[width=\textwidth,height=0.46\textheight,keepaspectratio]{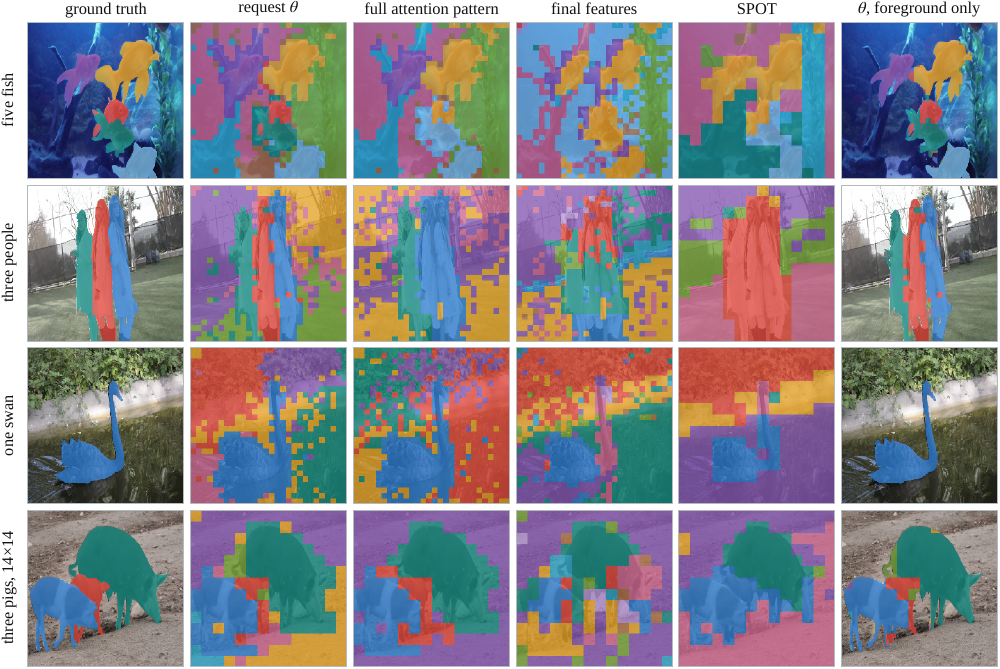}
  \captionof{figure}{\textbf{A useful consequence, and its boundary.} DAVIS frames grouped with the rule of Table~\ref{tab:coco} on a $28\times28$ grid ($14\times14$ for the pigs). SPOT is shown with its own $14\times14$ slots. Colors match corresponding regions across columns. The last column keeps only ground-truth foreground pixels, showing which errors come from including background. This column uses annotations and is not an automatic segmentation result.}
  \label{fig:grouping}
\end{minipage}\end{center}

\subsection{Reading from a region versus finding the patches of an object}

Direct attention shows where a query reads. Figure~\ref{fig:retrieval} also compares one source patch with every other patch through attention addresses, spatial requests and full attention patterns, all from DINOv2 L10h2. The source is a patch near the center of the largest annotated object. Direct attention is sparse, while the three similarity maps cover much of the source object and also spill into nearby background. Similar requests therefore mark patches that read from the same place, which is not the same as marking a complete object. This is why telling individuals apart and recovering object masks are different tests.

\begin{center}\begin{minipage}{\textwidth}
\centering
\includegraphics[width=.94\textwidth]{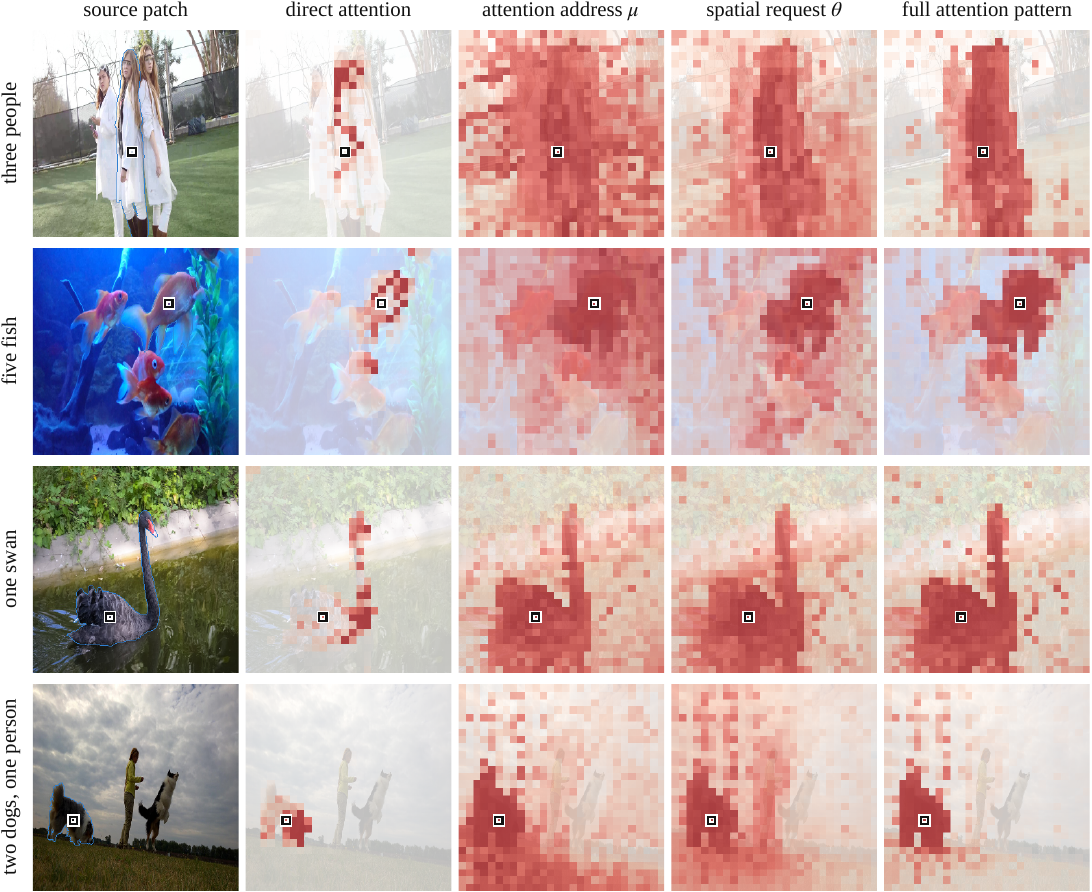}
\captionof{figure}{\textbf{The attention of one patch and its similarity to other patches.} Black squares mark the source. Only the first column uses its annotated outline. Higher scores appear darker red. For visibility, each panel uses its own minimum-to-99th-percentile contrast range, excluding the source when setting that range. Raw probabilities and similarity scores are not numerically comparable across columns. All maps use the same L10h2 head at $28\times28$ and show continuous patch values without spatial smoothing.}
\label{fig:retrieval}
\end{minipage}\end{center}

\newpage
\section{Further analyses}
\label{app:generality}

\subsection{Is the controlled test only about individuation?}

No. The controlled test removes category and position as cues, but what it finds is not specific to that setting. Keeping L10h2 and its spatial basis fixed, all four readouts also distinguish individuals when the negative object comes from another category (Table~\ref{tab:broaderbinding}). Request grouping likewise keeps its advantage across the most common object types (Figure~\ref{fig:categories}). The same routing therefore also separates objects of different categories.

\begin{center}\begin{minipage}{\textwidth}
\centering\small\setlength{\tabcolsep}{7pt}
\begin{tabular}{lcc}
\toprule
& \multicolumn{2}{c}{Different-instance negatives} \\
Readout & Same category & Different category \\
\midrule
Full attention pattern & .862 & .915 \\
Attention address $\mu$ & .867 & .850 \\
Spatial request $\theta$ & .866 & .845 \\
Requested destination $\widehat\mu$ & .858 & .811 \\
\bottomrule
\end{tabular}
\captionof{table}{\textbf{The frozen readouts distinguish individuals across categories too.} Mean same-instance AUC. At each shared offset, we sample equal numbers of same-instance, different-instance/same-category and different-category pairs, at most 12 of each. Each image contributes at least 12 pairs of each kind. All patches are foreground, and only the negative pair type changes between columns.}
\label{tab:broaderbinding}
\end{minipage}\end{center}

\subsection{How does this work differ from Li et al. (2025)?}
\label{app:libridge}

\citet{li2025binding} studied whether object binding emerges in pretrained ViTs, and decoded same-object membership with a learned probe. We study a more specific question: which native heads keep two objects of the same category apart, and how do they do it (Table~\ref{tab:licomparison})? The two tasks differ most in their negative pairs. In our reconstruction of their ADE20K protocol, only $1.0\%$ of pairs come from two instances of one category, so success on that task does not by itself show individuation. The readouts also differ. Their probe is trained on object labels, while our spatial fit uses none and describes a component of the native query that we can edit. On identical objects, their probe loses the separation toward the top of the network, while the late-layer heads of Appendix~\ref{app:copies} still keep identical copies apart.

\begin{center}\begin{minipage}{\textwidth}
\centering\small
\setlength{\tabcolsep}{6pt}
\renewcommand{\arraystretch}{1.12}
\begin{tabular}{>{\raggedright\arraybackslash}p{0.15\textwidth}>{\raggedright\arraybackslash}p{0.36\textwidth}>{\raggedright\arraybackslash}p{0.36\textwidth}}
\toprule
 & \citet{li2025binding} & Ours \\
\midrule
Question & Does object binding naturally emerge in pretrained ViTs? & How is same-category individuation expressed and used in native attention? \\
Problem & Binding: do two patches belong to the same object? & Individuation: which of two same-category objects does a patch belong to? \\
Pairs & All pairs among patches sampled within an image, including background and different categories & Foreground pairs, with negatives from another instance of the same category \\
Spatial control & No displacement matching & Match distance and direction exactly \\
Main readout & Learned quadratic probe & Native Q/K relations and the spatial request \\
Goal & Characterize binding geometry and effects of editing it & Identify specific layers and heads, then explain and edit their native routing \\
\bottomrule
\end{tabular}
\captionof{table}{\textbf{Two complementary questions.}}
\label{tab:licomparison}
\end{minipage}\end{center}

\subsection{Why does the controlled diagnostic matter?}

A broad same-object test rewards any readout that knows category or nearness, so it can point to the wrong heads. Figure~\ref{fig:taskcomparison} shows this for one query. With the pairs of \citet{li2025binding} on our VOC images, category agreement alone reaches $.948$ AUC, and the test selects L11h3 key similarity, which responds to both depicted buses. Once background pairs are removed, L10h2 attention rises to the top, but nearness still explains much of the score: nearby patches usually share an object. L9h7 key similarity shows this cue. It scores well while offsets are unmatched and drops once they are matched. With category and offset both controlled, L10h2 attention is still the selected head, and it reads from the bus of the query. The controls do not create the signal. They remove the easier cues, so that the selected head is the one that separates individuals.

\begin{center}\begin{minipage}{\textwidth}
\centering
\includegraphics[width=.98\textwidth]{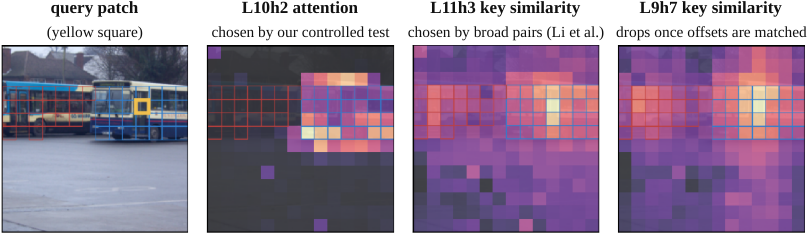}
\captionof{figure}{\textbf{Broad pair tests highlight different native readouts.} One query on the two-bus scene. L9h7 scores $.82$ AUC without offset matching and $.63$ with it. These maps illustrate the comparison, and the population results above establish it.}
\label{fig:taskcomparison}
\end{minipage}\end{center}

\subsection{Why keep the full request if the destination is two-dimensional?}
\label{app:faq_shape}

The requested destination says where a query reads. The full request also says how its preference spreads around that point: two requests can point to the same place while one favors a tight region and the other a wide one. For telling individuals apart, the destination is almost enough: on different parts of one person it reaches $.940$ AUC, against $.942$ for the full request. Curvature matters when we are interested in what attention does. With it, the requested destination predicts much better how the address moves under the request swap (Table~\ref{tab:curvature}), and the full request groups repeated objects better than its destination alone ($.615$ versus $.580$ FG-ARI). Cubic terms add no consistent gain. We keep the quadratic because it describes where and how a query reads, not because identity is five-dimensional.

\subsection{Is the selected head stable?}

Heads and layers are chosen on the VOC development images, then kept fixed on held-out VOC and COCO. Across 1,000 random re-splits of the 203 VOC images into development and held-out halves, the development rule selects L10h2 attention every time. Across backbones, the rankings of all heads correlate at $.93$--$.98$ between the two evaluation sets. In DINOv2, the selected head also has the strongest quadratic fit and request discrimination among all 144 heads (Figure~\ref{fig:lockedgeometry}). Related instance-sensitive heads occur in the other backbones, although their locations differ. The result concerns a recurring organization of attention, not one universal head index.

\clearpage
\section{Additional visual examples}
\label{app:visuals}

\begin{center}\begin{minipage}{\textwidth}
\centering
\includegraphics[width=\textwidth,height=.86\textheight,keepaspectratio]{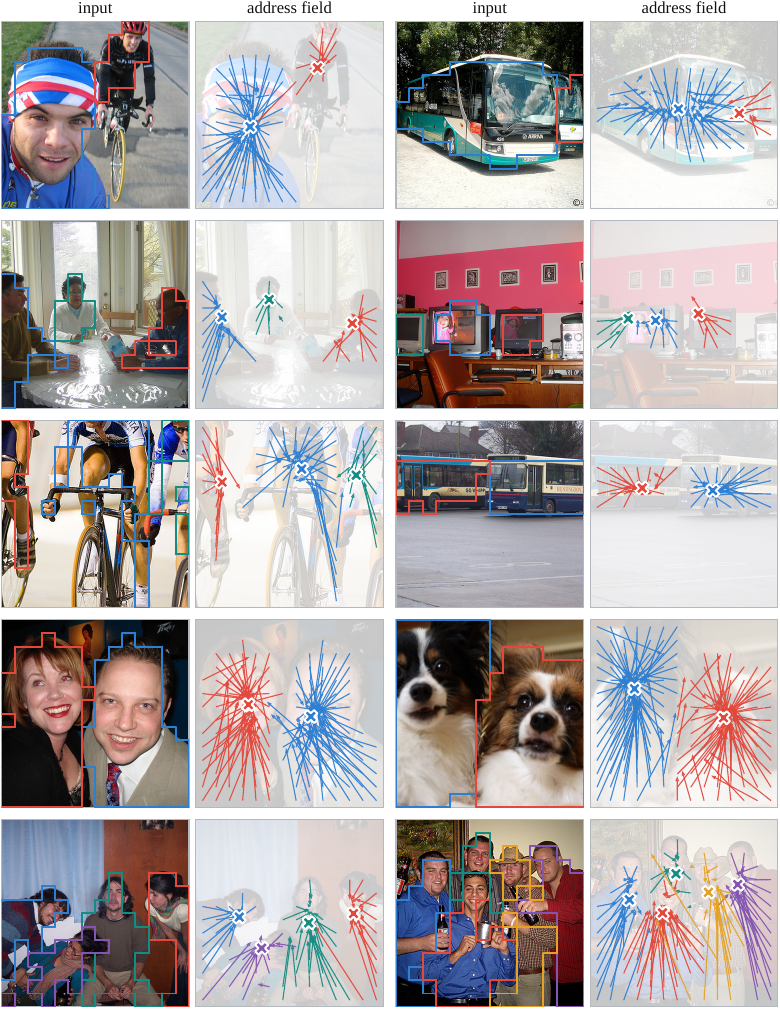}
\captionof{figure}{\textbf{One frozen DINOv2 head across held-out scenes.} Each pair shows the input and the L10h2 address field. Outlines identify the compared instances. The bottom row contains four and five people. These are illustrative scenes, not a quantitative cohort.}
\label{fig:address_examples}
\end{minipage}\end{center}
\begin{center}\begin{minipage}{\textwidth}
\centering
\includegraphics[width=\textwidth,height=.86\textheight,keepaspectratio]{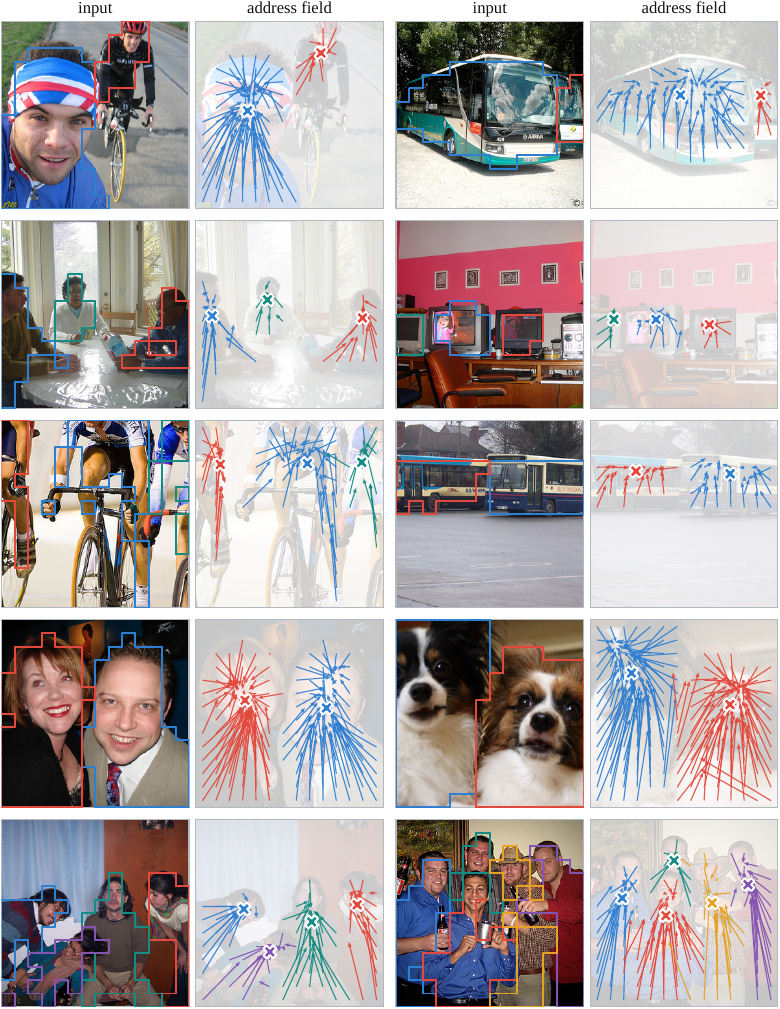}
\captionof{figure}{\textbf{The same scenes in DINOv3.} Address fields of L11h7. The quantitative comparison uses L10h1, the head selected on the development split. L11h7 is its neighbour, first of the 144 heads on held-out VOC and second on the development split. The bottom row adds four and five people, with each instance shown.}
\label{fig:address_examples_dinov3}
\end{minipage}\end{center}
\clearpage

\paragraph{How requests change with depth and neighboring objects.}
For the layer sequence of Figure \ref{fig:depth}, a separate spatial basis was fitted at each layer using the same 102 VOC development images, applied on one unchanged scene. It is a qualitative illustration, not another held-out estimate. The nearby-object examples (Fig. \ref{fig:overlap}) show that distinct requests can coexist even when instances touch or overlap. It does not imply that this head separates every touching pair.

\begin{center}\begin{minipage}{\textwidth}
\centering
\includegraphics[width=\textwidth]{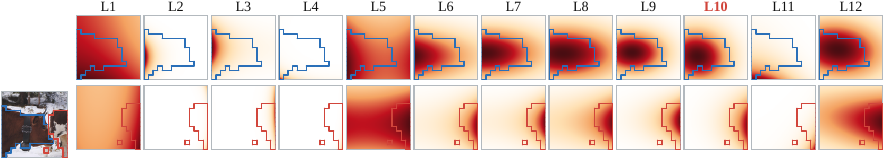}
\captionof{figure}{\textbf{Spatial requests through depth.} Mean request of each cow at every DINOv2 layer (head 2, with a spatial basis fitted separately per layer). In layers 2--4 both requests sit on the border. From layer 6 to 11 the two cows make distinct requests.}
\label{fig:depth}
\end{minipage}\end{center}
\begin{center}\begin{minipage}{\textwidth}
\centering
\includegraphics[width=.92\textwidth]{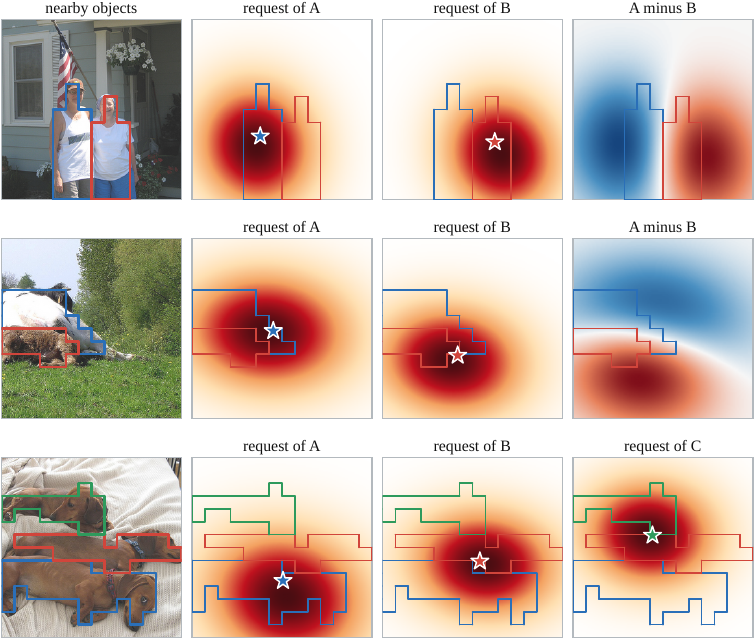}
\captionof{figure}{\textbf{Nearby instances make distinct requests.} Each row shows nearby instances, their mean requests and requested destinations ($\star$). The final column shows the request difference for two instances, or the request of the third dog.}
\label{fig:overlap}
\end{minipage}\end{center}

\end{document}